%% file: odor_arxiv.tex
\documentclass[preprint]{elsarticle}

\usepackage{lineno,hyperref}
\PassOptionsToPackage{unicode}{hyperref}
\modulolinenumbers[5]

\usepackage{amssymb}
\usepackage{subcaption}
\usepackage{xurl}
\usepackage[capitalize]{cleveref}
\usepackage{xspace}
\usepackage{microtype}
\usepackage[nolist]{acronym}
\usepackage{booktabs}
\usepackage{tikz}
\usepackage{pgfplots}
\pgfplotsset{compat=1.9}
\usepackage{multirow}
\usepackage{dirtytalk}
\usepackage{longtable}
\usepackage{multirow}

\usepackage{colortbl}
\usepackage{tabularx}
\usepackage{floatrow}
\definecolor{oorange}{HTML}{e69f00}
\definecolor{bblue}{HTML}{56b4e9}
\definecolor{ggreen}{HTML}{009e73}
\definecolor{yyellow}{HTML}{f0e442}
\definecolor{ppink}{HTML}{cc79a7}
\definecolor{DodgerBlue3}{HTML}{1874cd}
\definecolor{RedViolet}{HTML}{9e096d}
\definecolor{Firebrick3}{HTML}{cd2626}
\definecolor{forestgreen}{HTML}{556b2f}

\usepackage[inline]{enumitem}
\setlist*[enumerate]{label=(\arabic*)}

\journal{arXiv preprint}

\usepackage{siunitx}
\usetikzlibrary{backgrounds}

\usepackage{rebuttal}

\begin{document}

\begin{acronym}
    \acro{MADet}{Mutual-Assistance Learning for Object Detection}
    \acro{FCOS}{Fully Convolutional One-Stage Object Detection}
    \acro{DETR}{DEtection TRansformer}
    \acro{FocalNet}{Focal Modulation Network}
    \acro{CLIP}{Contrastive Language-Image Pre-Training}
    \acro{F-RCNN}{Faster Region Convolutional Neural Network}
    \acro{ILSVRC}{ImageNet Large Scale Visual Recognition Challenge}
    \acro{YOLO}{You Only Look Once}
\end{acronym}

\makeatletter
\DeclareRobustCommand\onedot{\futurelet\@let@token\@onedot}
\newcommand{\@onedot}{\ifx\@let@token.\else.\null\fi\xspace}
\makeatother
\newcommand{\etal}{\textit{et~al}\onedot}
\newcommand{\ie}{i.\,e.,\xspace}
\newcommand{\eg}{e.\,g.,\xspace}
\newcommand{\vs}{vs\onedot}
\newcommand{\aka}{a.\,k.\,a\onedot}
\newcommand{\cf}{cf\onedot}

\begin{frontmatter}

    \title{Smelly, Dense, and Spreaded: The \textit{O}bject \textit{D}etection for \textit{O}lfactory \textit{R}eferences (ODOR) Dataset}

\author[prl]{Mathias Zinnen\corref{cor}}
\ead{mathias.zinnen@fau.de}
\cortext[cor]{Corresponding Author}
\author[prl]{Prathmesh Madhu}
\ead{prathmesh.madhu@fau.de}
\author[knaw]{Inger Leemans}
\ead{i.b.leemans@vu.nl}
\author[mbg]{Peter Bell}
\ead{peter.bell@uni-marburg.de}
\author[prl]{Azhar Hussian}
\ead{azhar.hussian@fau.de}
\author[prl]{Hang Tran}
\ead{hang.tran@fau.de}
\author[knaw]{Ali Hürriyetoğlu}
\ead{ali.hurriyetoglu@dh.huc.knaw.nl}
\author[prl]{Andreas Maier}
\ead{andreas.maier@fau.de}
\author[prl]{Vincent Christlein}
\ead{vincent.christlein@fau.de}

\affiliation[prl]{organization={Pattern Recognition Lab, Friedrich-Alexander-Universität},%
            city={Erlangen},
            country={Germany}}
\affiliation[knaw]{organization={Royal Netherlands Academy of Arts \& Sciences (KNAW)},
            City={Amsterdam},
            country={Netherlands}}
\affiliation[mbg]{organization={Institut für Kunstgeschichte, Philipps-Universität},
            city={Marburg},
            country={Germany}}

\begin{abstract}
Real-world applications of computer vision in the humanities require algorithms to be robust against artistic abstraction, peripheral objects, and subtle differences between fine-grained target classes.
Existing datasets provide instance-level annotations on artworks but are generally biased towards the image centre and limited with regard to detailed object classes.
The proposed ODOR dataset fills this gap, offering 38,116 object-level annotations across 4,712 images, spanning an extensive set of 139 fine-grained categories.
Conducting a statistical analysis, we showcase challenging dataset properties, such as a detailed set of categories, dense and overlapping objects, and spatial distribution over the whole image canvas. 
Furthermore, we provide an extensive baseline analysis for object detection models and highlight the challenging properties of the dataset through a set of secondary studies. 
Inspiring further research on artwork object detection and broader visual cultural heritage studies, the dataset challenges researchers to explore the intersection of object recognition and smell perception. 
\end{abstract}

\begin{keyword}
Object Detection \sep Dataset \sep Computational Humanities \sep Olfaction \sep Small Object Detection \sep Transfer Learning
\PACS 42.30.Tz \sep 42.30.Sy
\MSC 68T45 \sep 1111
\end{keyword}

\end{frontmatter}

\section{Introduction}
Applying computer vision techniques to visual art images is more complex than applying them to real-world images. 
In contrast to artworks, real-world images, and even medical imaging modalities like X-Ray exhibit a certain amount of homogeneity in terms of style and image properties.
On the other hand, artistic representations of the same object can differ significantly depending on the degree of abstraction, artistic genre or the specific style and artistic abilities of the artist~\cite{cai2015cross}.
This leads to poor analysis and inferior performance of methods designed for real-world images when tested on artwork images~\cite{madhu2022enhancing}.

Artworks display varying degrees of abstraction ranging from naturalistic paintings that resemble photographs to abstract or nonrepresentational paintings having a complex understanding.
With recent progress made in the field of deep learning-based computer vision, large artwork datasets such as OmniArt~\cite{strezoski2018omniart} enable the learning of computer vision methods for image-level applications such as artist or genre classification.%
While significant progress has been made in the field of artwork image classification, there has been limited advancement in tasks such as object detection and segmentation for artworks and paintings. 
One of the primary reasons is the lack of publicly available annotated datasets in the artistic domain. 

\begin{figure}[t]
\begin{subfigure}{.33\columnwidth}
\centering
\includegraphics[width=.95\linewidth]{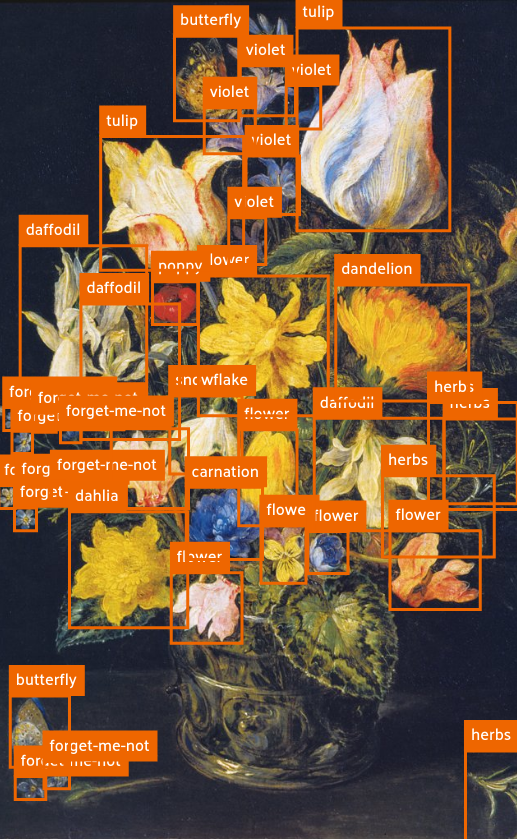}
\caption{}
\label{fig:examplesa}
\end{subfigure}\begin{subfigure}{.33\columnwidth}
\centering
\includegraphics[width=.95\linewidth]{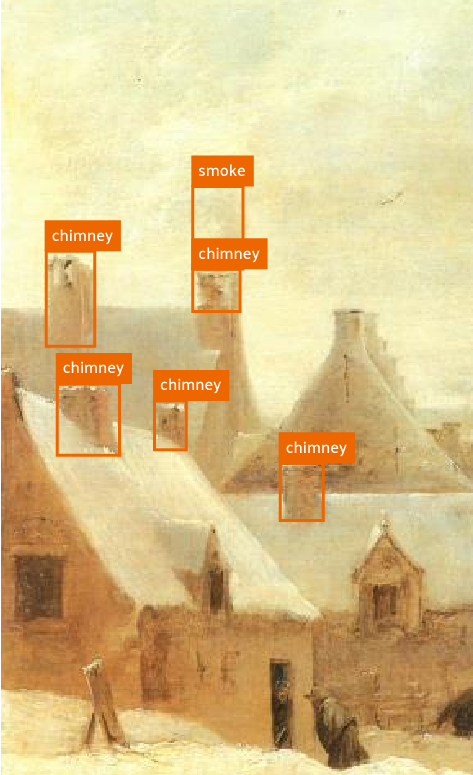}
\caption{}
\label{fig:examplesb}
\end{subfigure}\begin{subfigure}{.33\columnwidth}
\centering
\includegraphics[width=.95\linewidth]{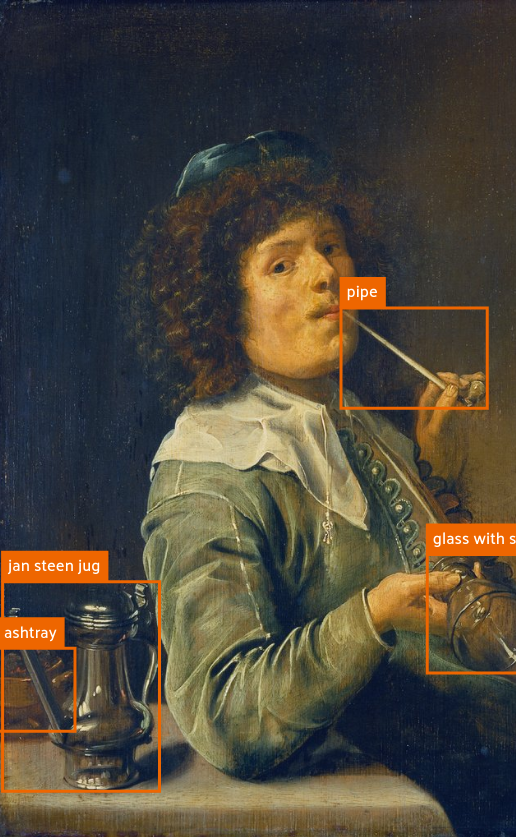}
\caption{}
\label{fig:examplesc}
\end{subfigure}
\caption{Examples of the ODOR dataset showing its variety with dense and overlapping instances (left), easy and large instances (middle), 
as well as occluded and ambiguous instances (right).
}
\label{fig:examples}
\end{figure}
In this work, we introduce a new medium-scale dataset that addresses two important research problems in understanding artworks: detecting a large number of categories and the precise 2-D localisation of small objects. 

Currently, very few datasets offer object-level annotations for artworks, and those available are limited in terms of object categories and complexity~\cite{westlake2016detecting, gonthier2018weakly, reshetnikov2022deart, zinnen2023sniffy, ju2023human}.
In contrast, the ODOR dataset stands out due to its complexity and features a detailed list of classes that extend beyond typical art-historical concepts.
Among existing detection datasets in the artistic domain, ODOR exhibits the greatest variability of object classes, resulting in a dense and complex spatial distribution across the entire image canvas, with 
objects often overlapping and displaying diverse sizes and aspect ratios (see \cref{fig:examples}). 
With its detailed class set and long-tailed distribution, ODOR can serve as a benchmark for object detection in low-sample regimes, comparable to the role of LVIS~\cite{gupta2019lvis} in the broader field of natural images.
Given the difficult properties of this dataset, we expect it to present significant new challenges.
This will promote the development of robust and capable detection algorithms for artworks and establish ODOR as one of the most challenging detection benchmark in the artwork domain.

A previous version of the dataset\footnote{\url{https://zenodo.org/record/7125961}} has already proven its value as a benchmark for a challenge about object detection in artworks~\cite{zinnen2022odor}.

Additionally, the dataset pushes research in the domains of digital humanities, especially digital olfactory heritage.
Emerging from the Odeuropa\footnote{\url{https:odeuropa.eu}} project, the dataset has been created in close collaboration with smell experts from multiple disciplines with the aim of enabling the automatic recognition of olfactory references. 
The dataset is grounded in a concrete research objective, leading to real-world properties like a fine granularity of detection categories and a diverse spatial distribution of the objects of interest.
These properties make it useful for many applications in digital heritage and render it a challenging dataset for object detection in non-natural domains.

\section{Related Work}
\input{related_work}

\section{The ODOR Dataset}
\input{odor_dataset}

\section{Benchmark Analysis}
\input{benchmark_analysis}

\section{Limitations}
\input{limitations}

\section{Conclusion}
We introduced the \textit{O}bject \textit{D}etection for \textit{O}lfactory \textit{R}eferences dataset (ODOR). 
It is the second-largest public artwork dataset with object-level annotations in terms of the number of images, and the most complex with regard to the number of classes and the density of annotated objects. 
With its help, the development of object detection algorithms that are robust against occlusion, small objects, and varying artistic representation can be advanced.
Additionally, its detailed class hierarchy may enable many applications in the domain of digital humanities, particularly in the developing field of digital sensory heritage.

We conducted extensive experiments with state-of-the-art object detection methods to understand how the characteristics of the proposed dataset affect their performance. An analysis of the performance improvement connected to stronger feature extraction backbones revealed that the dataset enables a high scope of learning robust representations of dense and complex objects. Interestingly, the YOLO architecture performs nearly as well as much heavier two-stage and transformer-based models on small objects.
This performance might be attributed to its anchorless region-proposal mechanism, which is well-suited for handling the complex challenges posed by the ODOR dataset. 
Another important factor is that the YOLO model benefits from pre-training on COCO, unlike other models that were only pre-trained for classification on ImageNet.
This suggests that exploring different pre-training schemes for gradual domain adaptation—from image-level classification to detection and from natural images to artworks—could be highly beneficial. 
The ODOR dataset thus offers a unique opportunity to experiment with these transfer learning methods. 
The insights gained from these experiments could extend beyond artworks and contribute to the broader field of general transfer learning.

In future releases of the dataset, we aim to further extend the number of images and annotations by applying a semi-automated strategy to annotate artworks from multiple digital museum collections. 
Furthermore, we plan to expand the amount of textual image-level metadata provided with the images to enable innovative multimodal and multitask approaches. 
We also aim to put the dataset to the test in future iterations of the ODOR challenge.
Overall, the ODOR dataset contributes to the field of digital humanities with its extensive annotations and diverse metadata, thus enabling the training of detection algorithms that extend beyond the typical focus of art history. 
Simultaneously, it establishes a challenging new benchmark that encourages the development of advanced algorithms better suited for artworks, beyond the typical focus on natural images.

\section*{Declaration of Competing Interest}
The authors declare that they have no known competing ﬁnancial interests or personal relationships that could have appeared to
inﬂuence the work reported in this paper.

\section*{Declaration of Generative AI and AI-assisted technologies in the writing process'}
During the preparation of this work, the authors used ChatGPT\footnote{\url{chat.openai.com}} in order to improve legibility and search for alternative phrasings. After using this tool/service, the authors reviewed and edited the content as needed and take full responsibility for the content of the publication.

\section*{Acknowledgements}
We thank the whole Odeuropa team for all the fruitful discussions, valuable feedback and patient explanations of how annotated visual data might be useful for quantitative approaches in the humanities.
We gratefully acknowledge the support of NVIDIA Corporation with the donation of the two Quadro RTX 8000 used for this research. 
This work was supported by the EU H2020 project under grant agreement No. 101004469.

\clearpage
 \bibliographystyle{elsarticle-num} 
 \bibliography{refs}
\clearpage
\appendix
\section{Image Credits}
\input{image_credits}
\pagebreak
\section{Experimental Details}
\input{training_details}

\pagebreak
\section{Author Biographies}
\input{authorbios}

\end{document}

%% file: related_work.tex
\paragraph{Computer Vision and the Humanities}
In the last decade, many computer vision tasks like object detection, pose estimation or image segmentation have been researched for real-world images.
Hence, they exhibit impressive performance on real-world and natural-looking images.
However, on evaluating artwork images, their performance is poor, leaving a large scope for improvement~\cite{zhao2023automatic, kadish2021improving}.
This can be attributed to the cross-depiction problem~\cite{cai2015cross,hall2015cross}, stating that the distance between features of artistic and natural representations of the same class is higher than that of different classes within their domain (natural/artistic)~\cite{cetinic2022understanding}.
Domain adaptation techniques like style transfer~\cite{kadish2021improving,lu2022data,madhu2022enhancing} or transfer learning \cite{sabatelli2018deep,gonthier2021analysis,zinnen2022transfer,zhao2022big} have been proposed to bridge this representational gap.
Another approach to address the challenges associated with the cross-depiction problem was proposed by Cheng \etal \cite{fisherdisc}, who apply a Fisher Discriminant Regularization to minimize within-class distances and maximize between-class distances in feature space.
Similar challenges found in the art domain also occur in other domains that diverge from the relatively standardized representation of natural scenes present in large-scale photographic datasets like ImageNet~\cite{russakovsky2015imagenet}, COCO~\cite{lin2014microsoft}, or OpenImages~\cite{kuznetsova2020open}. 
For instance, remote sensing images frequently suffer from large intra-class variances, occlusions, and varying object scales and aspect ratios~\cite{cheng2020remote}.
Techniques developed for remote sensing, like rotation-invariance~\cite{fisherdisc} or contextual feature enhancement~\cite{cheng2021feature}, may also prove beneficial when applied to artworks.

Although digital art history has become an established scholarly field, until recently, the employment of image processing and computer vision technologies has had limited applications in art history. 
Current developments in computer vision and deep learning techniques bring promising new angles for art historical research while branching out to other humanities domains.
This is illustrated by the significance of visual arts for the investigation of the past before the invention of photography in the 1820s:
Before the emergence of photographic evidence, artworks provide the main and often only visual representation of past epochs.
To leverage the wealth of knowledge encoded in historical artworks, various computer vision tasks like classification, object detection and scene understanding have been approached in digital humanities now. 
Crowley~\etal~\cite{crowley2014state,crowley2015search,crowley2016art} were the first to apply convolutional feature representations, learned from real-world images, to enable object recognition in paintings. 
One of the early available public datasets for object detection has been PeopleArt~\cite{westlake2016detecting}, where Westlake \etal applied a modified Fast R-CNN~\cite{girshick2015fast} to detect people in artworks. 
Later, Gonthier \etal~\cite{gonthier2018weakly} presented a weakly supervised learning approach, leveraging image-level labels to improve the object detection performance on a set of iconological artworks.
To further reduce the number of required labels, Madhu \etal~\cite{madhu2022one} apply one-shot learning to detect objects in images from Christian archaeology and art history.
For understanding artworks, pose estimation techniques were used to obtain new insights about the composition of artworks~\cite{madhu22icc, madhu2021understanding,bell2019ikonographie,bernasconi2022gab}.
Similarly, Impett \etal~\cite{impett2017totentanz,impett2020analyzing} cluster poses to analyse \mbox{Aby Warburg's concept of Pathosformeln~\cite{pathos}.}

In another strand of work, Kim \etal~\cite{kim2018seeing} and Eda \etal~\cite{eda2023detection} target the detection of smell-related objects specifically. 
However, unlike the ODOR dataset, these efforts focus on natural images, which limits their relevance to detecting smell-related objects in the visual arts.
In closer relation to the current work is the ODOR challenge~\cite{zinnen2022odor}, which focused on the detection of olfactory objects in artworks using a previous version of the dataset that included fewer categories and samples.

Identifying specific elements in large-scale image datasets with computer vision enables scholars to track similarities and changes over time, unveiling the production and reuse of visual patterns~\cite{rodriguez2020image, naslund2020cultures}.
These techniques can support the classical art historical research strands into authorship, motifs, visual styles, and compositions~\cite{lang2018reflecting}.
Moving away from the analysis of the image as a unique entity, digital object detection can help to research the `social life of things': the meaning of objects and the role they play in societies~\cite{appadurai1988social, hicks2010oxford}.
Tracing olfactory objects over time, genres and cultural domains can not only support new insights into the materiality of the objects and their cultural-historical significance, but it also supports a perception-based approach~\cite{van2021materials}. 
A seventeenth-century image of a peasant puffing out a mouth full of smoke from a clay pipe - a new consumer practice in Western Europe - surely triggered a sensory response in early modern viewers. What can we learn about these possible olfactory perceptions by following \say{scent trails} from depictions of smoking pipes, cigars, incense burners to other smokes and fumes, such as those of chimneys, gunfire, volcanoes and hell pits? 
Here, computer vision technologies also help to connect art history with the history of knowledge, researching images as instruments of meaning-making and knowledge production~\cite{leemans2022wind, rodriguez2020image}.

\paragraph{Object Detection Datasets}
The availability of large publicly available datasets with object-level annotations is one of the key factors in the success of deep learning for computer vision.
Since the release of the classic PASCAL VOC~\cite{everingham20062005} dataset, there has been an increasing research focus on releasing high-quality object detection datasets with millions of annotated images.
The most notable is the COCO dataset~\cite{lin2014microsoft}, which is widely used and still serves as the default benchmark for the evaluation of object detection algorithms. 
Recent real-world object detection datasets build upon COCO with different improvements.
For example, OpenImages~\cite{kuznetsova2020open} provides additional annotations for visual relations and a large number of 600 object classes. 
ImageNet also contains localisation information for millions of images across 1000 classes.
However, after the \ac{ILSVRC}~\cite{russakovsky2015imagenet}, it has not been widely adopted as a detection benchmark because the images typically feature only one large object in the image centre, which makes it much less challenging than COCO or other detection benchmarks. 
However, even these dedicated detection datasets exhibit a centre bias, indicating that the majority of annotated object instances are located in the middle of the image.
This can pose a challenge to detection models trained on these datasets, as they struggle with detecting objects close to the image borders~\cite{gupta2019lvis}.
Additionally, most of these datasets comprise real-world images, which combined with the cross-depiction problem makes it difficult to use them for learning artwork images.

\paragraph{Artwork Datasets}
The majority of the available datasets in the artistic domain are focused on image-level labels. 
Hence, most research tackles tasks like object recognition~\cite{crowley2015search}, artist or style classification~\cite{strezoski2018omniart}, image composition~\cite{madhu2023icc++}, or semantic tasks such as image description~\cite{garcia2018read} and visual question answering~\cite{garcia2020dataset}.
Very few artwork datasets~\cite{wilber2017bam,strezoski2018omniart} are comparable in size to real-world object detection datasets.

Apart from curated artwork datasets, numerous museum collections are slowly making their digitised collections publicly available. 
Meta-collections such as Europeana\footnote{\url{https://www.europeana.eu/de}}, RKD\footnote{\url{https://rkd.nl/en/}}, or the Iconclass explorer\footnote{\url{https://iconclass.org/}} provide gateways to multiple collections at once. 
For an extensive overview of galleries, libraries, and museums with open access to their collections, the reader is referred to~\cite{glamsurvey}. 

The metadata of a museum collection is usually confined to image-level information such as artist, material, and sometimes descriptions. 
There are only a few datasets available with object-level annotations, an overview of which is presented in \cref{tab: ds_overview}.
\begin{table}[t]
\centering
\caption{Overview of artwork datasets with object-level annotations, including the proposed ODOR dataset (b/img=boxes per image, cls/img=classes per image).}
\label{tab: ds_overview}
\begin{tabular}{lccccc}
\toprule
    Name & images & boxes & classes & b/img & cls/img \\
     \midrule
     PeopleArt~\cite{westlake2016detecting} &                       \phantom{0}1644&  \phantom{00}3870 &\phantom{00}1 & 2.4 & \phantom{0.}1 \\
     IconArt~\cite{gonthier2018weakly} &                            \phantom{0}1480 & \phantom{00}4931 & \phantom{0}10 & 3.3 & 1.7 \\
     DEArt~\cite{reshetnikov2022deart}\textsuperscript{$\dagger$} & 15157 &          111891 & \phantom{0}70 & 7.8 & 3.0 \\
     SniffyArt~\cite{zinnen2023sniffy}\textsuperscript{$\dagger$} & \phantom{00}441 & \phantom{00}1941 & \phantom{1}   & 4.4 & \phantom{0.}1 \\
     PoPArt\textsuperscript{$\ddagger$}~\cite{schneider2023poses} & \phantom{0}2454 & \phantom{0}10749 & \phantom{00}1 & 4.4 & \phantom{0.}1 \\
     Human-Art\textsuperscript{$\ddagger$}~\cite{ju2023human} & 50000 & 123000 & \phantom{00}1 & 2.5 & \phantom{0.}1 \\
     \textbf{ODOR} (ours)                                          &\phantom{0}4712 & \phantom{0}38116 &           139 & 8.1 & 3.7 \\
     \bottomrule   
     \multicolumn{6}{l}{\footnotesize \textsuperscript{$\ddagger$}Person annotations include keypoints.}\\
     \multicolumn{6}{l}{\footnotesize \textsuperscript{$\dagger$}Person annotations include gesture/pose labels.}
\end{tabular}
\end{table}
Westlake \etal introduced PeopleArt~\cite{westlake2016detecting}, a dataset consisting of 1644 artworks where the persons are annotated with bounding boxes.
The artworks are categorised into 41 diverse artistic movements with significantly different object representations, \eg cubism or realism.  
Gonthier \etal~\cite{gonthier2021analysis} provide object-level annotations only for a test set of 1480 images for 6 classes, whereas the full dataset contains about 6000 iconographic images.
In the second version of their dataset~\cite{gonthier_nicolas_2018_4737435} they extended the number of classes to 10 while leaving the set of images unchanged.
Recently, the release of DEArt~\cite{reshetnikov2022deart} has marked a large leap in both the number of annotated images and the diversity of categories. 
Using a semi-automatic approach, the authors were able to create \num{118981} annotations across 70 categories based on COCO which were then modified according to their historical date of appearance and art-historical significance~\cite{marinescu2020improving}.
Additionally, they provide pose classifications for all human-like objects, \eg angels, nude, or knights.
Similarly, the recently published PoPArt~\cite{schneider2023poses} and Human-Art~\cite{ju2023human} datasets focus on persons, providing bounding boxes and additional key point annotations.
Specifically for olfactory artwork understanding, the SniffyArt dataset~\cite{zinnen2023sniffy} provides a combination of person boxes, pose key points, and smell gesture labels to enable the automatic recognition of smell gestures.

Recent instance-level artwork datasets attempt to close the quantitative gap with real-world datasets. 
They contribute immensely concerning dataset size and the number of annotations. 
However, these datasets still fall short when compared to real-world counterparts like OpenImages (with 600 categories) or Objects365~\cite{shao2019objects365} (with 365 categories) in terms of fine-grained categorization.
Their large number of classes and their real-world properties remain unmatched by any artwork dataset.

Our proposed ODOR dataset fills this gap by providing object annotations for a diverse set of 139 fine-grained categories.
The object annotations in the proposed dataset exhibit challenging real-world properties. 
The spatial distribution of objects covers the whole image canvas removing the centre-biased property of available datasets.
Additionally, the dataset comprises many small objects in close proximity that often occlude each other, introducing density and occlusion as challenging characteristics of the dataset.

%% file: odor_dataset.tex
\subsection{Distribution Format}
\begin{figure}
    \centering
    \includegraphics[width=\textwidth]{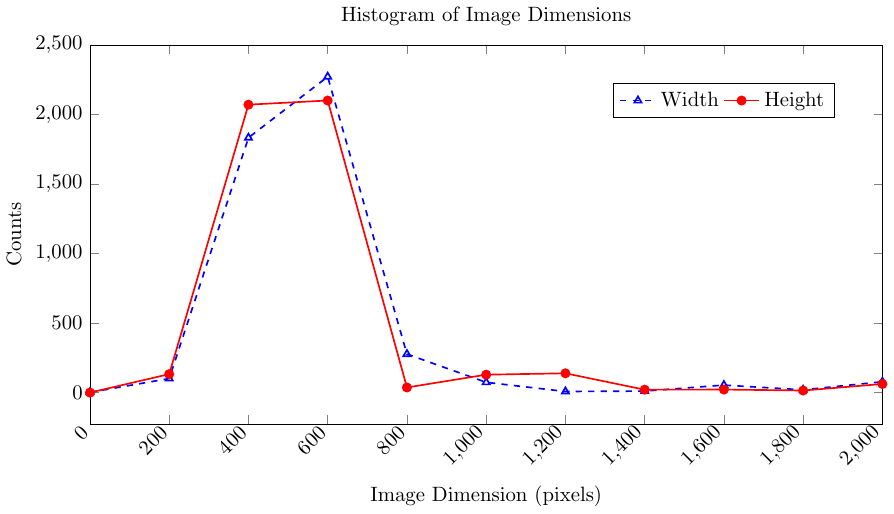}
    \caption{Distribution of image widths (blue, dashed) and heights (red) of the artworks in the ODOR dataset. Note that sizes above 2000 pixels in each dimension are integrated into the rightmost histogram bin to increase legibility.}
    \label{fig:imsizes}
\end{figure}

The annotations are provided in COCO JSON format.\footnote{\url{https://COCOdataset.org/\#format-data}}. 
To represent the two-level hierarchy of the object classes, we use the $supercategory$ field in the $categories$ array as defined by COCO.
In addition to the object-level annotations, we provide an additional CSV file with image-level metadata, which includes content-related fields, such as Iconclass codes~\cite{couprie1978iconclass,brandhorst2016iconclass}) or image descriptions, as well as formal annotations, such as artist, license, or creation year. 
The images have varying sizes ranging from 100x650 pixels to 3292x4000 pixels, depending on the resolution provided by their source collections. 
On average they have a width of 653 and a width of 641 pixels.
\cref{fig:imsizes} illustrates the distribution of image widths and heights for all images in the dataset.

Besides physical copies of the images for download on Zenodo,\footnote{\url{https://zenodo.org/record/11070878}} we enable their download from the respective source collections with links in the metadata file and a download script via Zenodo\footnote{\url{https://zenodo.org/record/11070878}}, GitHub\footnote{\url{https://github.com/mathiaszinnen/odor-dataset}}, and HuggingFace\footnote{\url{https://huggingface.co/datasets/mathiaszinnen/odor}}.

\subsection{Dataset Acquisition}
The ODOR dataset has been created in the context of the interdisciplinary Odeuropa project\footnote{\url{https://odeuropa.eu}}, which aims to reconstruct and promote the European olfactory heritage by automatically analysing large amounts of visual and textual data. 
This is reflected in the focus of the image collection, which was to find artworks that relate to smell and olfaction.
To gather the artworks, we queried multiple digitized museum collections.\footnote{\url{https://rkd.nl},\url{http://catalogo.fondazionezeri.unibo.it},\url{https://bildindex.de},\url{https://pop.culture.gouve.fr},\url{https://sammlung.staedelmuseum.de},\url{https://wga.hu},\url{https://slub-dresden.de},\url{https://boijmans.nl},\url{https://nga.gov}}
As our knowledge about contexts in which smell-active objects might appear evolved during the dataset creation, we iteratively extended the image base with new search terms. 
This iterative extension was performed in close collaboration with smell experts from multiple disciplines such as art history, olfactory sciences, and museology.

\begin{figure*}[ht]
    \centering
    \includegraphics[width=\linewidth]{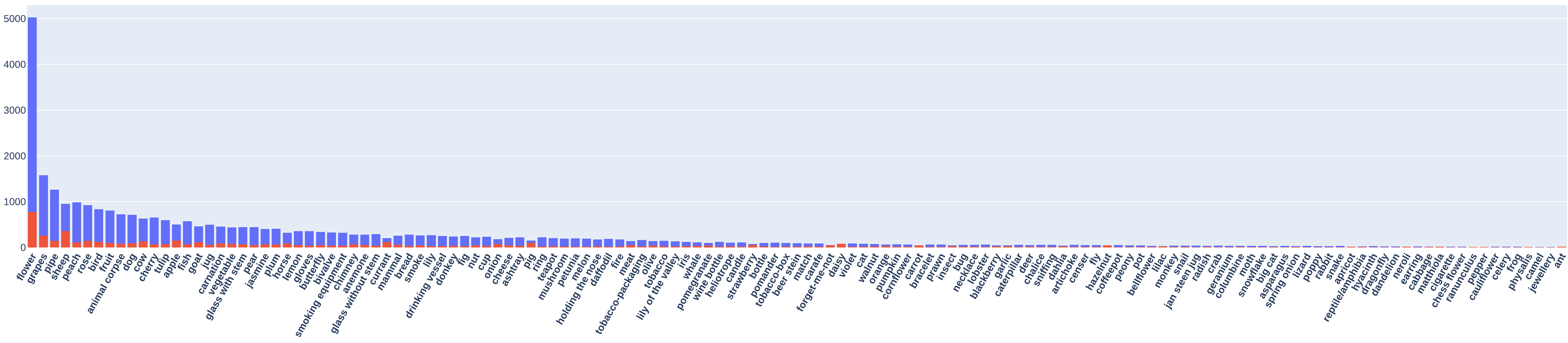}
\caption{Class distribution plot for the ODOR dataset train (blue) and test (red) splits; the long-tailed distribution of the rare classes poses a challenge for detection algorithms. Best viewed digitally.}
    \label{fig:classdist}
\end{figure*}

Similarly, a predefined set of categories was initially considered and then iteratively extended, eventually resulting in a set of 227 categories. 
This set of smell-relevant object categories is based on controlled vocabularies, which are a part of the Odeuropa data model~\cite{lisena2022capturing}. %
The large number of object categories, including very rare and special objects, suggests the use of a hierarchical class structure.
This approach has multiple advantages: 
\begin{enumerate*}
    \item It facilitates finding specific object categories, thus simplifying the annotation process.
    \item Detection systems can resolve to a fallback solution in cases where the exact object category cannot be determined but a broader classification can be made, \eg detecting a flower instead of its species. 
\end{enumerate*}

In contrast to the WordNet\cite{miller1995wordnet}-based hierarchy applied by Redmon \etal~\cite{redmon2017yolo9000}, we incorporate only two levels of abstraction for the sake of comprehensibility.
When assigning supercategories, we favour pragmatic considerations such as the visual similarity of subcategories or simplicity of hierarchy over strict taxonomic consistency.
This leads to a biologically incorrect hierarchy, \eg whales being classified as fishes instead of mammals.
However, it allows detection algorithms to predict the coarse category ``fish-like animals" based on visual (and olfactory) similarities rather than learning a biological taxonomy. %
\Cref{tab:supercategories} lists the resulting supercategories, along with their subcategories and corresponding instances in the ODOR dataset.
\begin{table}[t]
    \caption{A list of supercategories, number of respective subclasses and instances in the ODOR dataset. ``Other'' includes 12 supercategories with less than 1000 instances.}
    \centering
    \begin{tabular}{lcr}
    \toprule
        Supercategory & No.~ Subclasses  & No.~ Instances \\
        \midrule
        flower & 30 & 10866 \\
        fruit & 17 & 7818 \\
        vegetable & 15 & 1657\\
        mammal & 14 & 4972 \\
        drinking vessel & 14 & 2821 \\
        smoking equipment & \phantom{0}8 & 2595 \\
        smoke-related & \phantom{0}5 & 1031\\
        other & 36 & 6356 \\
        \bottomrule
    \end{tabular}
    \label{tab:supercategories}
\end{table}
The annotations were generated using a combination of crowd annotation using the Amazon Mechanical Turk (AMT) platform and manual labelling by expert annotators with a background in art history.
All crowd annotations have manually been checked by experts and multiple rounds of corrections were performed, especially in cases where expert knowledge was required, \eg for the accurate classification of specific flower species or historical drinking vessels (as depicted in \cref{fig:examplesa} and \cref{fig:examplesc}).

\subsection{Image-Level Metadata}
\begin{figure}
\begin{subfigure}{\textwidth}
    \centering
    \includegraphics[width=\textwidth]{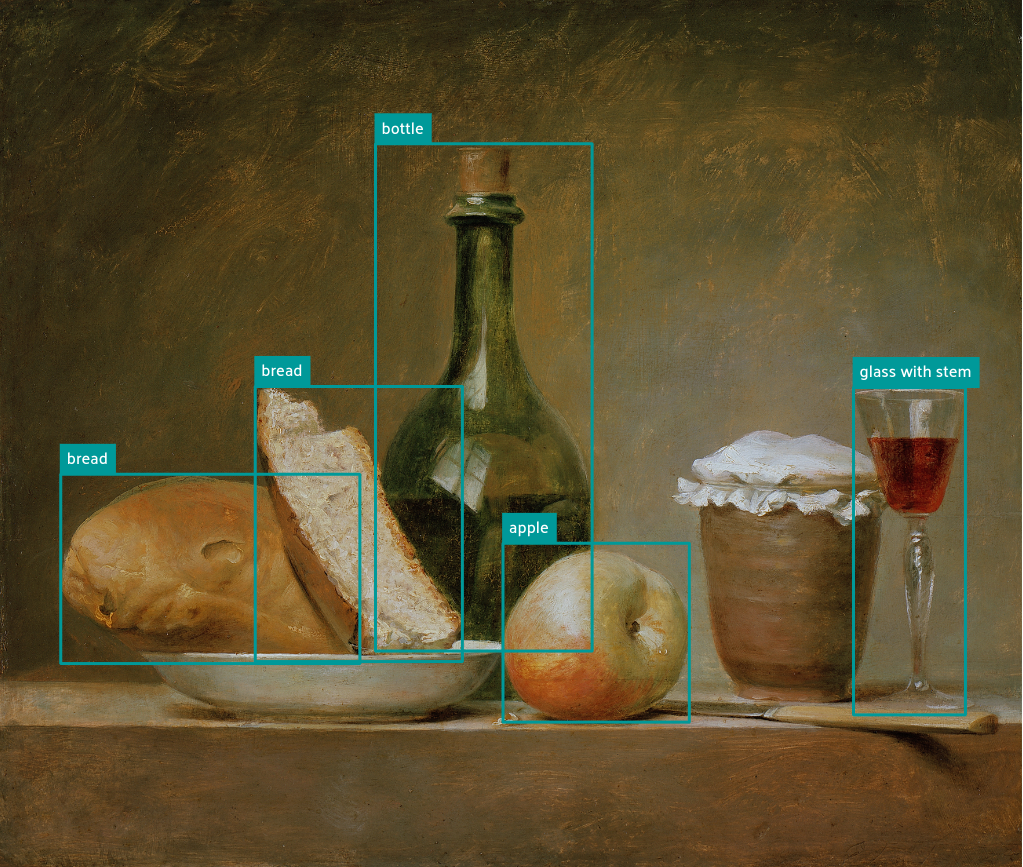}
    \captionsetup{labelformat=empty}
    \caption{Streng pyramidal sind die Gegenstände vor dem dunklen Fond arrangiert. Ein Teller mit geteiltem Brotlaib vermittelt zur aufragenden bauchigen Flasche, auf der sich das Atelierfenster der Künstlerin spiegelt. Ein gedachtes Dreieck wird durch Apfel, Konfitürentopf und das schmale, mit Rotwein gefüllte Glas beschrieben. Der rechts über die Steinkonsole ragende Messergriff unterbricht geschickt die Tektonik des Bildes. Treffend wurden die kontrastierenden Materialien geschildert. 
Der Malerin waren natürlich die Werke ihres Pariser Zeitgenossen Chardin vertraut. Im Gegensatz zu dessen atmosphärischem Kolorit verwendete Anne Vallayer-Coster leuchtendere Töne, die auf dekorative Wirkung zielten. Als Mitglied der Königlichen Kunstakademie unterhielt die Künstlerin im Louvre ein Atelier. Sie vollendete die glanzvolle Tradition der französischen Stilllebenmalerei des 18. Jahrhunderts.}
\end{subfigure}
\caption{Artwork from the ODOR dataset with rich image description.}
\label{fig:description}
\end{figure}

\begin{figure}
\begin{subfigure}{\textwidth}
    \centering 
    \includegraphics[width=.65\textwidth]{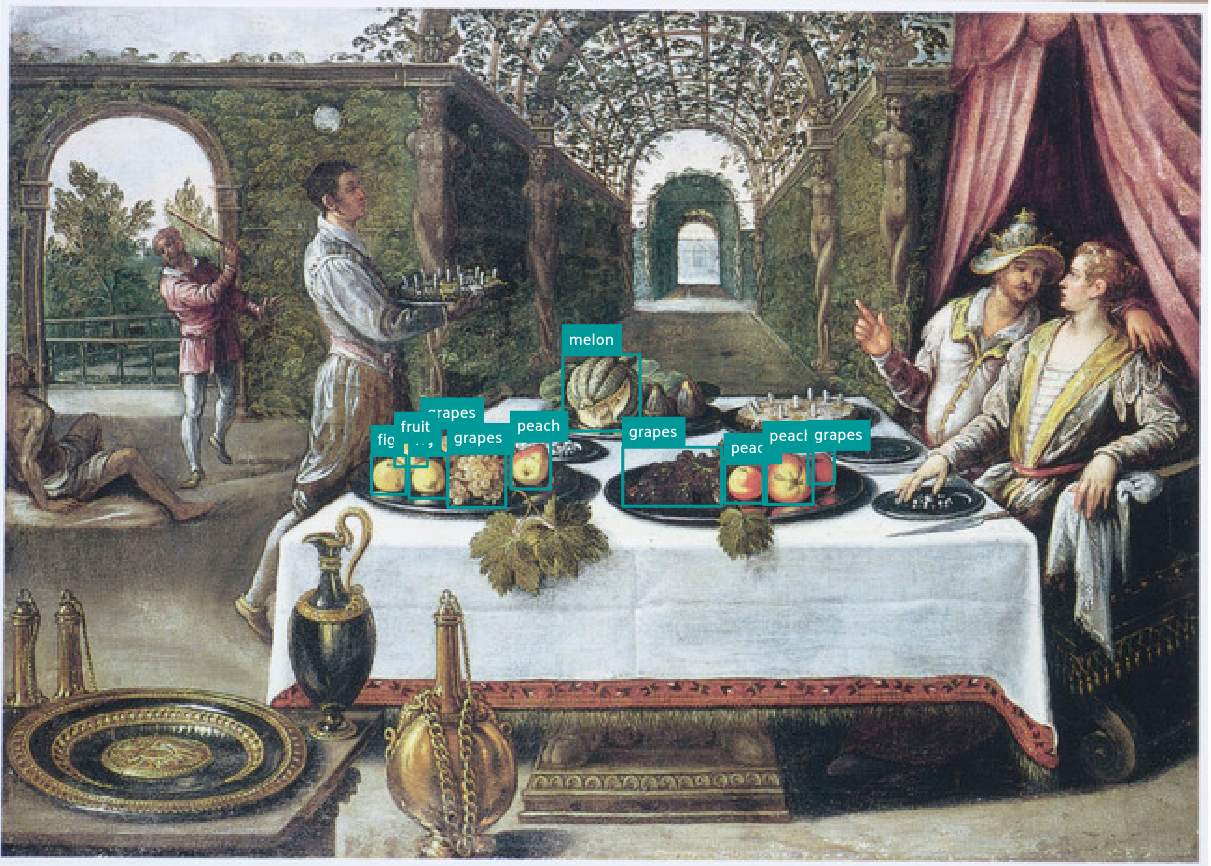}
\end{subfigure}
\smallskip

\begin{subfigure}{\textwidth}
\centering
\scriptsize{
    \begin{tabular}{ll}
        \toprule 
        Title & The parable of the rich man and poor Lazarus \\
        Iconclass Code & 73C852 \\
        Keywords & New Testament and Apocrypha, meal, garden \\
        \bottomrule
    \end{tabular} 
}
\end{subfigure}
\caption{Exemplary metadata for an artwork from the dataset. For brevity, only selected metadata fields are displayed, please refer to \cref{credits} for more metadata.}
\label{fig:metadata}
\end{figure}

Recent advances in multimodal image processing, particularly based on \ac{CLIP}~\cite{radford2021learning}, have demonstrated the potential of leveraging textual metadata information for detection tasks~\cite{kamath_mdetr_2021, li_grounded_2022, liu_grounding_2023, xie_described_2023}.
Although not the primary focus of this work, we included the metadata gathered from the various source collections while collecting the images. 
These collections do not share a common metadata representation; therefore, we consolidated the metadata into a unified schema as described in \cref{tab:metadata}.
\begin{table}[t]
    \centering
    \begin{tabularx}{\textwidth}{lXr}
    \toprule
     Metadata Field & Description & Coverage  \\
     \midrule
         File Name & Unique name of the image file & \SI{100}{\percent} \\
         Artist & Creator of the artwork & \SI{97.7}{\percent} \\
         Title & Artwork title & \SI{99.4}{\percent}\\
         Earliest Date & Earliest assumed date of artwork creation & \SI{99.2}{\percent}\\
         Latest Date & Latest assumed date of artwork creation& \SI{94.3}{\percent} \\
         Iconclass Code & Classification according to the Iconclass~\cite{couprie1978iconclass} scheme & \SI{12.6}{\percent}\\
         Description & Free form description of the image contents and/or providing additional information & \SI{3.8}{\percent}\\
         Keywords & Keywords describing the image content, standardized within collections & \SI{81.8}{\percent}\\
         License & License according to source collection & \SI{27}{\percent}\\
         Query Term & Search term used to query the image from the source collections & \SI{82.8}{\percent}\\
         Current Location & Location of the artwork at the time of retrieval & \SI{74.1}{\percent}\\
         Repository Number & Repository number in the source collection & \SI{93.1}{\percent}\\
         Photo Archive & URL of the source collection & \SI{100}{\percent}\\
         Image Credits & Image download link from the source collection & \SI{100}{\percent}\\
         Details URL & Link to the image in the context of its source collection & \SI{100}{\percent}\\
         \bottomrule
    \end{tabularx}
    \caption{Overview of image-level metadata fields in the ODOR dataset and a short description. Coverage denotes the percentage of images where the respective metadata has a nonempty value.}
    \label{tab:metadata}
\end{table}
Note, however, that not all metadata fields were present in each collection, resulting in a sparse coverage of many fields.
This is particularly true for the \say{description} field, which offers a valuable perspective for multimodal approaches. 
Especially for this property, the dataset could benefit from future enhancements, possibly including synthetic metadata.
\Cref{fig:description} showcases one of the relatively rare examples from the dataset, where the image description contains rich information about the artwork contents and its creation context.
\cref{fig:metadata} shows another example with an iconclass code, an instructive title and image-level keywords from the RKD collection.
The metadata presents opportunities to experiment with innovative algorithms that integrate multiple types of supervision signals. 
For instance, the descriptive image titles could be used to fine-tune a CLIP~\cite{radford2021learning}-based model on the ODOR dataset for enhanced detection, to incorporate additional textual information into detection algorithms, or to learn related tasks like artist classification in a multi-task setting.

\subsection{Dataset Statistics}\label{subsec:datastats}

\begin{figure}
    \centering
        \begin{tikzpicture}[inner frame xsep=0]
          \begin{axis}[
        			xlabel near ticks,
        			xlabel={Number of categories},
        			ylabel={Image percentage},
              height=6cm, width=\textwidth,
        			ytick style={draw=none},
                    ymax = .4,
                    ytick = {0,0.1,0.2,0.3,0.4},
                    yticklabels={0,10, 20,30,40},
        			xtick={1,2,...,15},
        			ymajorgrids = true,
        	    xtick pos=left,
        			enlarge x limits=0.02,
            ]
            \addplot[draw=oorange, mark=o, mark size=1pt,thick] %
            table[col sep=comma,x=x,y=y]{eval/cats_COCO.csv};
            \addplot[draw=ggreen, mark=o, mark size=1pt, thick] %
            table[col sep=comma,x=x,y=y]{eval/cats_IconArt.csv};
            \addplot[draw=bblue, mark=o, mark size=1pt, thick] %
            table[col sep=comma,x=x,y=y]{eval/cats_DEArt.csv}; 
            \addplot[draw=red, mark=o, mark size=1pt, thick] %
            table[col sep=comma,x=x,y=y]{eval/cats_ODOR.csv};
            \addlegendentry{COCO}
            \addlegendentry{IconArt}
            \addlegendentry{DEArt}
            \addlegendentry{ODOR}
          \end{axis}
        \end{tikzpicture}
        \begin{tikzpicture}[inner frame sep=0]
          \begin{axis}[
        			xlabel near ticks,
        			xlabel={Number of instances\phantom{g}},
        			ylabel={Image percentage},
                    ymax = 0.4,
                    ytick = {0,0.1,0.2,0.3,0.4},
                    yticklabels={0,10, 20,30,40},
        			xtick={1,2,...,15},
              height=6cm, width=\textwidth,
        			ymajorgrids = true,
        	    xtick pos=left,
        			enlarge x limits=0.02,
           legend pos=north east
            ]
            \addplot[draw=oorange, mark=o, mark size=1pt, thick] %
            table[col sep=comma,x=x,y=y]{eval/inst_COCO.csv};
            \addplot[draw=ggreen, mark=o, mark size=1pt, thick] %
            table[col sep=comma,x=x,y=y]{eval/inst_IconArt.csv};
            \addplot[draw=bblue, mark=o, mark size=1pt, thick] %
            table[col sep=comma,x=x,y=y]{eval/inst_DEArt.csv}; 
            \addplot[draw=red, mark=o, mark size=1pt, thick] %
            table[col sep=comma,x=x,y=y]{eval/inst_ODOR.csv};

          \end{axis}
        \end{tikzpicture}
    \caption{Number of annotated categories (top), and instances (bottom) per image for DEArt, IconArt, and ODOR.
    PeopleArt is not included since it has only one category.} 
    \label{fig:pgfpercentages}
\end{figure}
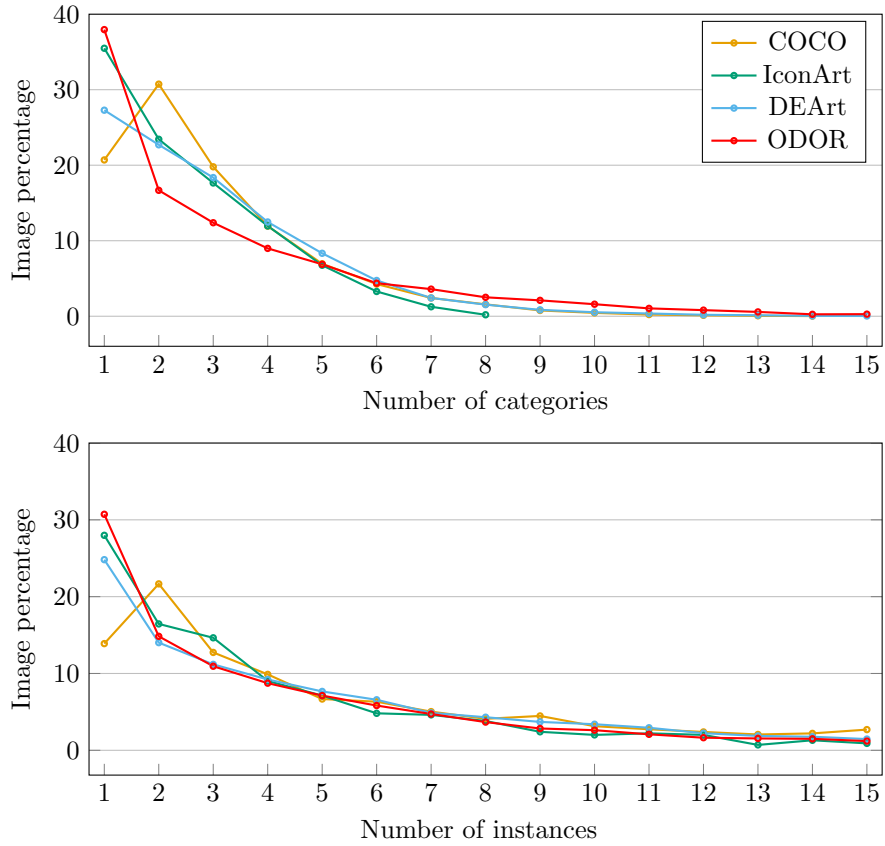

\paragraph{Basic Statistics}
A summary of the datasets representing the number of object categories and the number of instances per category is presented in \cref{tab: ds_overview}. 
The proposed dataset has a significantly higher number of object instances and categories than PeopleArt\cite{westlake2016detecting}, IconArt\cite{gonthier2018weakly} and PopArt\cite{schneider2023poses}.
On average, our dataset contains 3.7 categories and 8.1 instances per image. 
In comparison, PeopleArt\cite{westlake2016detecting}, IconArt\cite{gonthier2018weakly}, and PoPArt\cite{schneider2023poses} have less than 2 categories and 4.5 instances per image on average. 
ODOR has fewer images than DEArt\cite{reshetnikov2022deart}, but more categories and instances per category, which presents a new challenge for object detection. 
We summarise the number of categories and instances per image in \cref{fig:pgfpercentages}. 
Similar to Lin \etal~\cite{lin2014microsoft}, we argue that the images in our dataset have a lot of contextual information due to many categories and instances per image.

\begin{figure}
    \centering
    \includegraphics[width=\columnwidth]{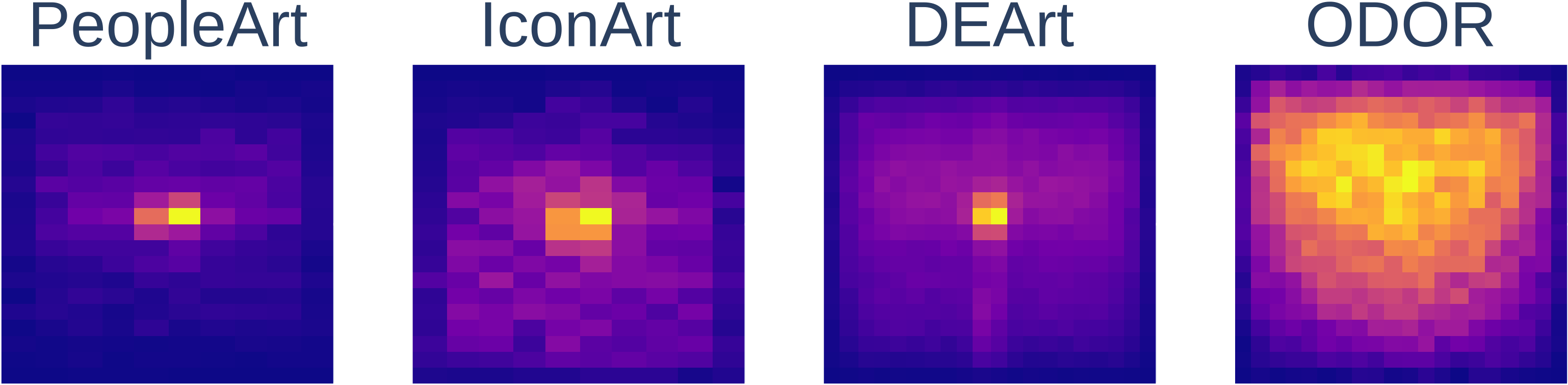}
    \caption{Spatial distribution of object centres in normalised image coordinates of various datasets.}
    \label{fig:spatial}
\end{figure}

\paragraph{Spatial Distribution}
The spatial distribution of objects in COCO\cite{lin2014microsoft} and PASCAL VOC\cite{everingham2009pascal} are biased towards the image centre, which can decrease detection performance for objects that are close to the border~\cite{pont2015boosting}.
More recent object detection datasets, such as OpenImages\cite{kuznetsova2020open}, ADE20\cite{zhou2017scene}, or LVIS\cite{gupta2019lvis}, aim for greater spatial diversity but are still biased towards the centre~\cite{gupta2019lvis}.

We plot the spatial distribution of the object centres for different artwork datasets in \cref{fig:spatial}.
In comparison to other datasets, the objects of our ODOR dataset are spread out much more evenly across the entire image canvas. 
This might be related to the much larger number of categories, capturing less significant objects that might be considered as background from an art-historical perspective.
This is in contrast to the other artwork datasets, where the fewer classes mostly cover objects that are in the artists' attention, usually in the centre of the image.

\begin{figure}[t]
    \centering
        \begin{tikzpicture}
          \begin{axis}[
        			xlabel near ticks,
        			xlabel={Overlapping instances},
        			ylabel={Instance percentage},
              height=6cm, width=\textwidth,
        			ytick style={draw=none},
                    ymax = .5,
                    ytick = {0,0.1,0.2,0.3,0.4,.5},
                    yticklabels={0,10, 20,30,40,50},
        			xtick={0,1,2,...,10},
                    xmax=10,
        			ymajorgrids = true,
        	    xtick pos=left,
        			enlarge x limits=0.02,
           legend pos=north east      
            ]
            \addplot[draw=oorange, mark=o, mark size=1pt,thick] %
            table[col sep=comma,x=x,y=y]{eval/peopleart_n_inst.csv};
            \addplot[draw=ggreen, mark=o, mark size=1pt, thick] %
                table[col sep=comma,x=x,y=y]{eval/iconart_n_inst.csv};
            \addplot[draw=bblue, mark=o, mark size=1pt, thick] %
            table[col sep=comma,x=x,y=y]{eval/deart_n_inst.csv}; 
            \addplot[draw=red, mark=o, mark size=1pt, thick] %
            table[col sep=comma,x=x,y=y]{eval/odor_n_inst.csv};
            \addplot[draw=gray, mark=o, mark size=1pt, thick] %
            table[col sep=comma,x=x,y=y]{eval/coco_n_inst.csv};
           \addlegendentry{Peopleart}
            \addlegendentry{IconArt}
            \addlegendentry{DEArt}
            \addlegendentry{ODOR}
            \addlegendentry{COCO}
          \end{axis}
        \end{tikzpicture}
        \begin{tikzpicture}
          \begin{axis}[
        			xlabel near ticks,
        			xlabel={Area of overlap\phantom{g}},
                    ymax = 0.5,
                    xmax=1,
              height=6cm, width=\textwidth,
        			ylabel={Instance percentage},
                    ytick = {0,0.1,0.2,0.3,0.4,.5},
                    yticklabels={0,10, 20,30,40,50},
        			ymajorgrids = true,
        	    xtick pos=left,
        			enlarge x limits=0.02,
            ]
            \addplot[draw=oorange, mark=o, mark size=1pt, thick] %
            table[col sep=comma,x=x,y=y]{eval/peopleart_area.csv};
            \addplot[draw=ggreen, mark=o, mark size=1pt, thick] %
            table[col sep=comma,x=x,y=y]{eval/iconart_area.csv};
            \addplot[draw=bblue, mark=o, mark size=1pt, thick] %
            table[col sep=comma,x=x,y=y]{eval/deart_area.csv}; 
            \addplot[draw=red, mark=o, mark size=1pt, thick] %
            table[col sep=comma,x=x,y=y]{eval/odor_area.csv};
            \addplot[draw=gray, mark=o, mark size=1pt, thick] %
            table[col sep=comma,x=x,y=y]{eval/coco_area.csv};
 
          \end{axis}
        \end{tikzpicture}
    \caption{Percentage of instances per number of overlapping instances (top), and area of overlap (bottom) for PeopleArt~\cite{westlake2016detecting}, DEArt~\cite{reshetnikov2022deart}, IconArt~\cite{gonthier_nicolas_2018_4737435}, COCO~\cite{lin2014microsoft} and ODOR datasets.
    } 
    \label{fig:pgfoccl}
\end{figure}
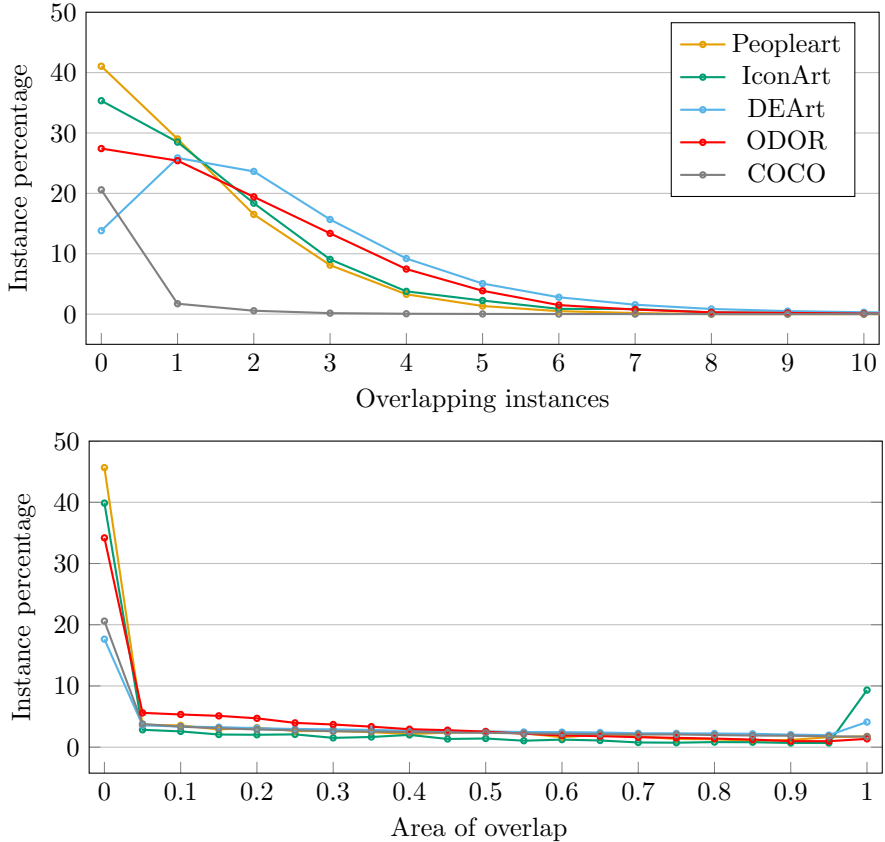

\paragraph{Occlusion}
In \cref{fig:pgfoccl}, we quantify the amount of occlusion in the proposed dataset by measuring \begin{enumerate*}
\item \textit{instance density}: the proportion of instances per number of overlapping instances (top), and
\item \textit{instance area overlap}: the proportion of instances per instance area shared with other instances (bottom).
\end{enumerate*}

\begin{figure}[t]
\begin{subfigure}{.5\columnwidth}
    \centering
    \includegraphics[width=.98\linewidth]{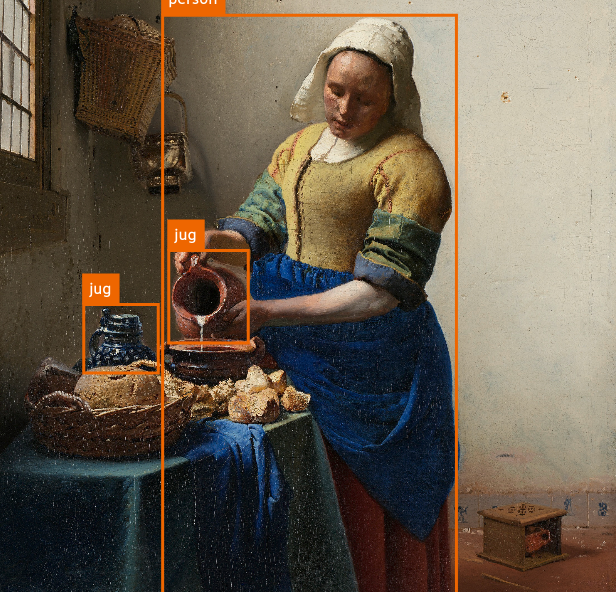}
    \caption{}
    \label{fig:occla}
\end{subfigure}~%
\begin{subfigure}{.5\columnwidth}
    \centering
    \includegraphics[width=.98\linewidth]{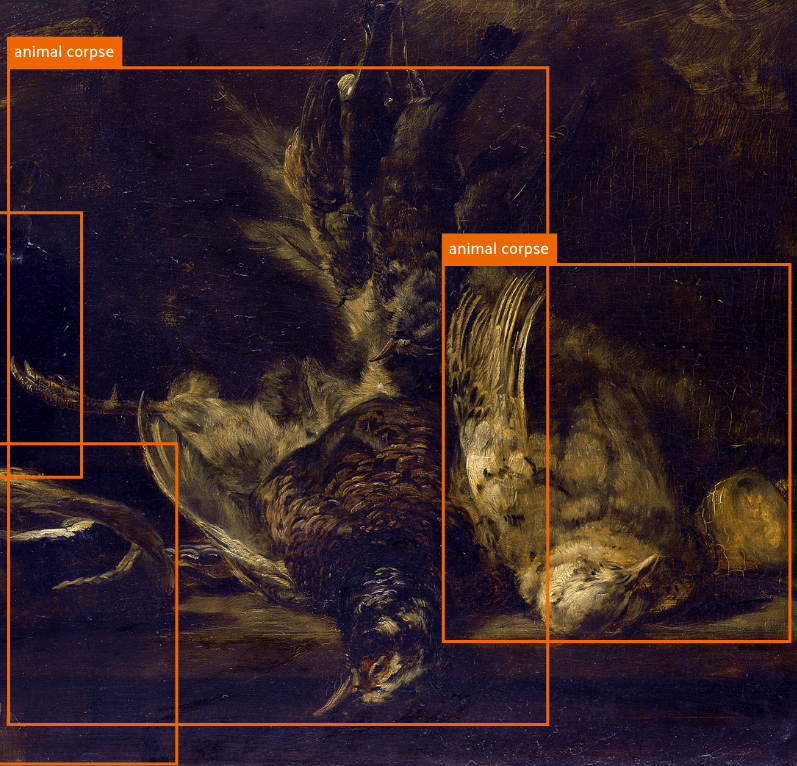}
    \caption{}
    \label{fig:occlb}
\end{subfigure}%
    \caption{Examples from the DEArt~\cite{reshetnikov2022deart} dataset (left) and the ODOR dataset (right) displaying varying amounts of occlusion.}
    \label{fig:occl}
\end{figure}

DEArt~\cite{reshetnikov2022deart} and ODOR share a high \textit{instance density} with more than one overlapping instance. 
Another possible way of quantifying the amount of occlusion in a dataset is by measuring the \textit{instance area overlap}.
However, we have to interpret these numbers carefully. 
Consider the \textit{depiction of a milkmaid} from the DEArt\cite{reshetnikov2022deart} dataset depicted in \cref{fig:occl} (left). 
One of the ``jug'' instances is located completely within the bounding box of the surrounding ``person'' instance and hence has a 100\% area of overlap. 
The dead birds depicted in \cref{fig:occl} (right), on the other hand, have only a small overlap of ${\sim}10\%$. 
Quantitatively, we would expect an instance with 100\% overlap to be much harder to detect than instances with smaller amounts of overlapping areas.
Subjectively, however, it is much easier to localise both jug instances in the left image than to distinguish between the only slightly occluding instances in the right image. 
Judging from the amount of overlap alone, we cannot make assumptions about the overall dataset complexity or detection difficulty.

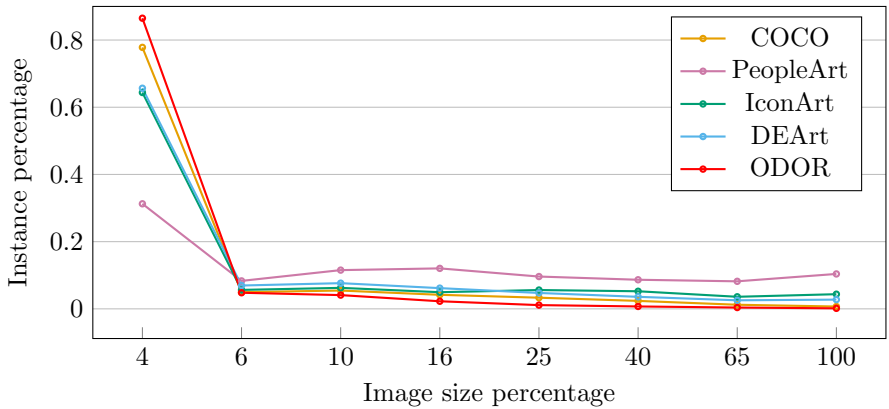
\begin{figure}[h]
    \centering
    \begin{tikzpicture}
          \begin{axis}[
        			xlabel near ticks,
        			xlabel={Image size percentage},
                    xmin = 0.5,
                    xmax = 8.5,
                    xtick={1,2,...,8},
                    xticklabels={4,6,10,16,25,40,65,100},
                    ymax = .9,
        			ylabel={Instance percentage},
                    height=6cm, width=\textwidth,
        			ytick style={draw=none},
        			ymajorgrids = true,
        	        xtick pos=left,
                    legend pos = north east
            ]
            \addplot[draw=oorange, mark=o, mark size=1pt, thick] %
            table[col sep=comma,x=x,y=y]{eval/sizes_COCO.csv}; 
            \addplot[draw=ppink, mark=o, mark size=1pt,  thick] %
            table[col sep=comma,x=x,y=y]{eval/sizes_PeopleArt.csv};
            \addplot[draw=ggreen, mark=o, mark size=1pt, thick] %
            table[col sep=comma,x=x,y=y]{eval/sizes_IconArt.csv}; 
            \addplot[draw=bblue, mark=o, mark size=1pt,  thick] %
            table[col sep=comma,x=x,y=y]{eval/sizes_DEArt.csv}; 
            \addplot[draw=red, mark=o, mark size=1pt,  thick] %
            table[col sep=comma,x=x,y=y]{eval/sizes_ODOR.csv};
            \addlegendentry{COCO}
            \addlegendentry{PeopleArt}
            \addlegendentry{IconArt}
            \addlegendentry{DEArt}
            \addlegendentry{ODOR}
          \end{axis}
    \end{tikzpicture}
    \caption{Histogram of instance box sizes, relative to the image size.}
    \label{fig:pgfsizes}
\end{figure}

\paragraph{Object Sizes}
Due to the limited amount of information encoded in their pixel representation small objects pose a distinct challenge for object detection methods and require more contextual reasoning to recognise~\cite{liu2021survey}. 
\Cref{fig:pgfsizes} gives a histogram of the bounding box size relative to the image size.
Compared to other artwork datasets the ODOR dataset has a high proportion of very small objects covering less than 4\% of the image area while it has only very few large objects covering more than 25\%. 

\subsection{Dataset Splits}

The full set of annotations obtained during the data acquisition phase consists of \num{38249} object instances across \num{227} categories in \num{4732} images.
To guarantee a fair evaluation as well as a minimum amount of training data, we only retain categories with at least \num{16} instances distributed over at least three separate images.
This way, we ensure that each category can be represented at least once in a training, validation, and test split. 
The heuristic choice of \num{16} minimum instances results from balancing the goal of maximising the number of instances and categories with the requirement to provide enough instances for training and evaluation.

The instances of the eliminated categories are either removed from the dataset or assigned to their respective supercategory, depending on whether the class hierarchy permits such an assignment.
The resulting final ODOR dataset consists of \num{38116} object instances in \num{4712} images across \num{139} categories.
We split it into train and test sets according to a 90:10 ratio, resulting in a train set comprising \num{4264} images and \num{32251} bounding boxes while the test set consists of \num{448} images and \num{5865} bounding boxes.
In addition to the train and test splits, we will release unfiltered annotations with the full set of 227 categories.

The class distribution of the ODOR dataset is illustrated in \cref{fig:classdist}.
The long tail distribution poses a challenge for detection algorithms, which have to learn to recognise rare classes from only a few samples of training data.
We aim to promote the development of models for more generic object detection in artworks similar to how LVIS~\cite{gupta2019lvis} did for object detection and instance segmentation in natural images. 
A further challenge is posed by the fact that the relatively small objects in the ODOR dataset are occluding each other. 
\Cref{fig:occl} gives two representative examples of artworks with a large number of small and occluding objects.

\subsection{Dataset Applicability}
Although the selection of images and object classes is centred around olfactory relevance, applications of the dataset are not restricted to this specific area.

Naturally, its modest size in comparison to large-scale photographic datasets limits the dataset's suitability for training detection algorithms targeting natural images.
However, it can assume a valuable role as a complementary benchmark when the COCO performances of algorithms are very close or in scenarios necessitating robustness against challenges like occlusion, small objects, or off-centre objects. 
Furthermore, it can be used for training and evaluation in applications requiring broad generalisation across multiple domains or styles.

As a testament to the dataset's value as a benchmark for object detection within the artistic sphere stands the ODOR challenge~\cite{zinnen2022odor}, where the participants were challenged to detect a set of smell-active objects in an earlier dataset iteration.

In computational humanities, the extensive set of object categories allows for a broader range of applications than comparable art-centric datasets would permit. 
An illustration of this is provided by van Erp \etal~\cite{van2023more}, where the prevalence of roses in still-life artworks across four centuries was analysed quantitatively.
With its focus on common \say{background-like} objects the dataset prompts a perceptual shift away from the centre scene which traditionally has been the focal point of art history.
Instead, it directs our attention to the outer edges of the images, encompassing plants, animals, foodstuffs, and vessels, but also less delineated \say{objects} such as smoke or fire.
This aligns with the paradigm of the \say{material turn} in the humanities~\cite{hicks2010oxford}, thus opening new lines of investigation, \eg by considering the \say{social life of things}~\cite{appadurai1988social} or enabling quantitative approaches in microhistory~\cite{magnusson2013microhistory}

Specifically in the niche (but rapidly developing) field of sensory and olfactory heritage, the dataset serves as the first resource to enable the automated analysis of visual smell references.
Previous versions of the dataset have already been used in multiple olfaction-related applications, including various challenges~\cite{zinnen2022odor, ali2022musti}, semantic web knowledge integration~\cite{lisena2022capturing}, or quantitative approaches to smell history~\cite{van2023more}.
The increasing significance of olfaction in the field of humanities and cultural heritage is evidenced by a growing interest in multi-sensual approaches, particularly in relation to smell~\cite{howes2010sensual,tullett2022smell}.

%% file: benchmark_analysis.tex
We demonstrate the challenging properties of the ODOR dataset by applying various state-of-the-art object detection methods. 
The aim is to gain valuable insights into the performance of various families of object detection methods when trained to detect a large number of spatially distributed and partially occluding object classes on artwork images.
Code to reproduce the baseline methods can be found on GitHub\footnote{\url{https://github.com/mathiaszinnen/odor-dataset}}.

Following common practice, we initialize all feature extraction backbones with pre-trained weights on ImageNet. 
Depending on the model size, we use either the original ImageNet-1k~\cite{russakovsky2015imagenet} weights, pre-trained for classifying 1000 categories (ResNets~\cite{he2016deep}, ResNexts~\cite{xie2017aggregated}, SWIN-T~\cite{liu2021swin}), or ImageNet-21k~\cite{ridnik2021imagenet} weights, which were trained to distinguish among 21,000 classes (SWIN-L~\cite{liu2021swin}, FocalNet~\cite{yang2022focal}).

\subsection{Methods}

We evaluate three different families of object detection methods: \begin{enumerate*}
   \item a classical two-stage anchor-based approach,
   \item a state-of-the-art transformer-based architecture,
   \item three one-stage detection methods.
\end{enumerate*}

\paragraph{Two-Stage Approach}
  \ac{F-RCNN}~\cite{ren2015faster} is the dominant representative of the classical two-stage object detection method.
   The main idea is to apply a region proposal network to generate a set of proposal predictions and then refine and classify them in a second step. 
    Due to various improvements~\cite{he2017mask,lin2017feature,cai2018cascade} over the last decade, it is still considered a standard approach for object detection.
    For many detection tasks, using one of the many implementations and fine-tuning it on a small amount of custom data is the easiest solution. 
    Hence, we chose F-RCNN~\cite{ren2015faster} with multiple established backbones, including ResNet~\cite{he2016deep}, ResNeXt~\cite{xie2017aggregated}, and the more recent SWIN transformer~\cite{liu2021swin} as the baseline benchmark.
    \paragraph{DETR-Based Architecture} Many of the best-performing object detection algorithms (according to the COCO test-dev benchmark~\cite{paperswithcode}) use the DINO~\cite{zhang2022dino} architecture with varying backbones~\cite{yang2022focal,wang2022internimage}. 
    DINO is based on the \ac{DETR}~\cite{carion2020end} architecture and improves upon it by applying a contrastive denoising training loss, and an innovative query selection for anchor initialisation.
        We evaluate DINO~\cite{zhang2022dino} with a strong SWIN~\cite{liu2021swin} transformer backbone, and the recently published \ac{FocalNet}~\cite{yang2022focal}.
    Currently, \ac{FocalNet}~\cite{yang2022focal} is among the best-performing object detection methods with available code~\cite{paperswithcode}. 
    \paragraph{One-Stage Methods} 
   One-stage methods eliminate the need for a separate region proposal stage as they directly predict boxes and their classification from the input image. 
    Typically, this results in faster training and inference times at the cost of a relatively weaker performance compared to the more complex two-stage or transformer-based methods.
    Here we evaluate the performance of \ac{FCOS}~\cite{tian2019}, \ac{MADet}~\cite{xie2023}, and YOLO-v8~\cite{Jocher_YOLO_by_Ultralytics_2023}.
    Both \ac{FCOS}~\cite{tian2019} is and YOLO-v8~\cite{Jocher_YOLO_by_Ultralytics_2023} are anchor-free methods.
    Unlike F-RCNN~\cite{ren2015faster}, they do not require a predefined set of anchor boxes per proposal location but instead directly generate predictions for each point of the feature maps.
    This gives them more flexibility to handle uncommon object sizes and aspect ratios~\cite{tian2019,xie2023}.
    It is particularly interesting to investigate how well the anchorless region proposal mechanism implemented in \ac{FCOS}~\cite{tian2019} and YOLO-v8~\cite{Jocher_YOLO_by_Ultralytics_2023} performs against the large number of small and complex objects in the ODOR dataset.
    \ac{MADet}~\cite{xie2023}, on the other hand, merges the anchor-based and anchor-free paradigms by defining two branches that align anchor-based and anchor-free bounding box predictions.

All models are initialised with pre-trained weights before fine-tuning them on our dataset. 
We use the default parameter settings of each model for training ODOR models and evaluate them against the standard metrics used in the COCO benchmark. 
In particular, we list the mean average precision (AP) computed over different intersection over union (IoU) cut-off thresholds (50 to 95 with a step of 5), the average precision with IoU threshold 50 (AP$_{50}$), and the average precision with IoU threshold 75 (AP$_{75}$).

\subsection{Results}

\begin{table}
     \caption{Comparison of the performance of representative detector configurations on the ODOR test set when varying the detection heads (\subref{tab:results_arch}) and the feature extraction backbone (\subref{tab:results_backbone}).}
    \label{tab:results_new}
\begin{subtable}{.48\textwidth}
    \centering
    \begin{tabular}{lcc}
              \toprule
            Detector &   AP & AP$_{50}$ \\
             \midrule
             F-RCNN\cite{ren2015faster}  &  10.1 & 19.3  \\
              DINO\cite{zhang2022dino}  &  \textbf{13.3} & \textbf{22.2}  \\
              MADet\cite{xie2023} &   10.2 & 18.7  \\
              FCOS\cite{tian2019}  &  8.0 & 14.3  \\
             \bottomrule
    \end{tabular}
     \caption{Fixed ResNet-50 Backbone \color{RedViolet}}
     \label{tab:results_arch}
\end{subtable}%
\begin{subtable}{.48\textwidth}
    \begin{tabular}{llcc}
              \toprule
            Backbone & PT DS &  AP & AP$_{50}$ \\
             \midrule
            RN50\cite{he2016deep} & IN1k & 10.1 & 19.3 \\
            RN101\cite{he2016deep} & IN1k & 10.6 & 19.4 \\
            x101-64\cite{xie2017aggregated} & IN1k & 11.9 & 21.5 \\
            SWIN-L\cite{liu2021swin} & IN21k & \textbf{19.0} & \textbf{35.1} \\
            \bottomrule
\end{tabular}
\caption{Fixed F-RCNN Detector \color{RedViolet}}
\label{tab:results_backbone}
\end{subtable}
\end{table}

\paragraph{Model Performance}

\begin{figure*}
\centering
\begin{subfigure}{.24\linewidth}
\centering
\includegraphics[width=.95\linewidth]{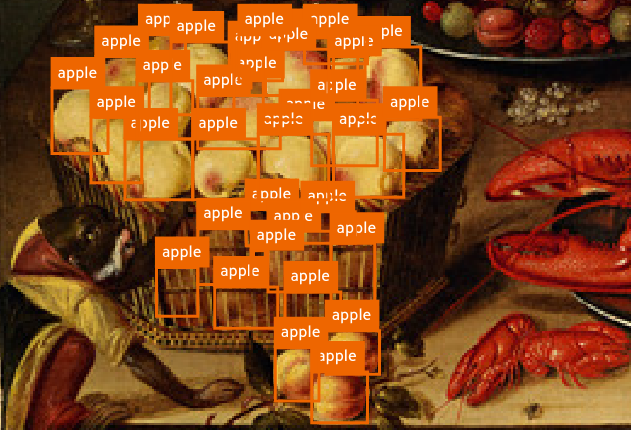}
\caption{Ground Truth}
\label{fig:gt}
\end{subfigure}\begin{subfigure}{.24\linewidth}
\centering
\includegraphics[width=.95\linewidth]{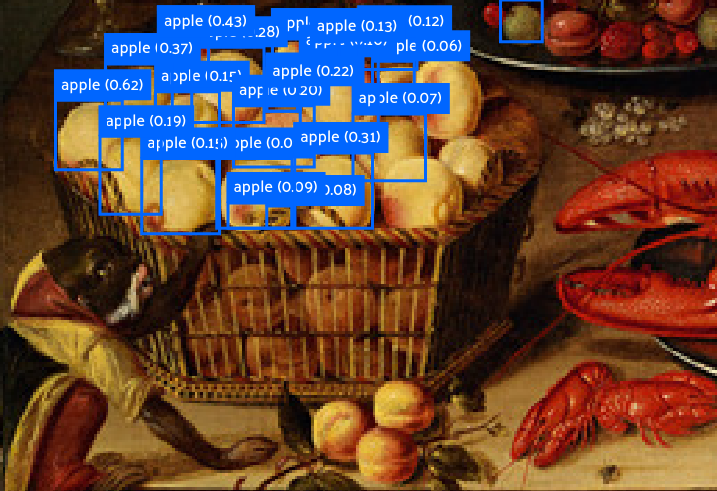}
\caption{DINO~(FocalNet-L)}
\label{fig:focaldino}
\end{subfigure}
\begin{subfigure}{.24\linewidth}
\centering
\includegraphics[width=.95\linewidth]{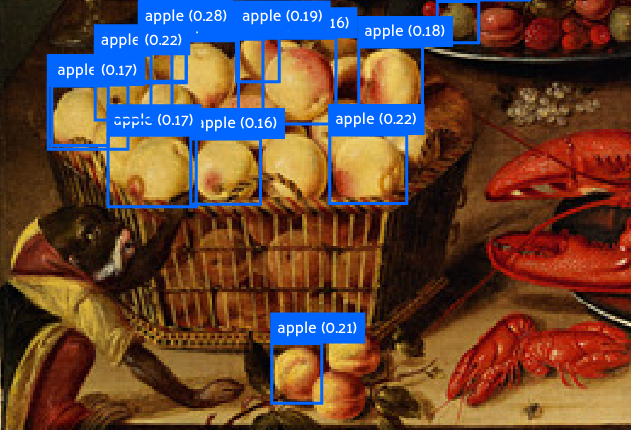}
\caption{DINO~(SWIN-L)}
\label{fig:dinoswinl}
\end{subfigure}\begin{subfigure}{.24\linewidth}
\centering
\includegraphics[width=.95\linewidth]{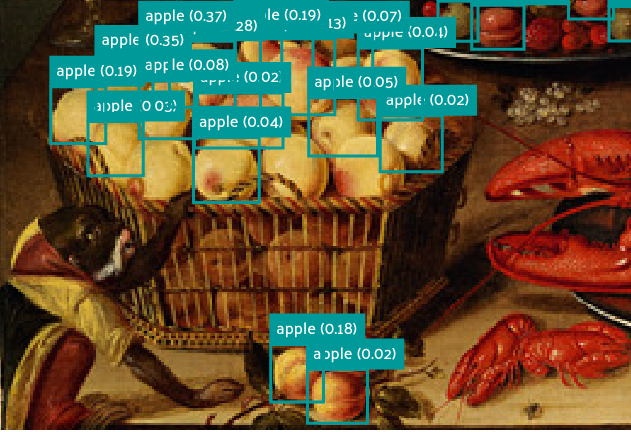}
\caption{YOLO-v8L}
\label{fig:yolov8preds}
\end{subfigure}
    \caption[]{Example detections on a heap of small, partly occluding apples. Zoomed in for illustrative purposes. We observe that the anchorless YOLO-v8 model \subref{fig:yolov8preds} outperforms the otherwise stronger DINO models with \subref{fig:focaldino} heavy FocalNet-L and \subref{fig:dinoswinl} SWIN-L backbones. Ground truth boxes are shown in \subref{fig:gt}. For clarity, only ``apple'' class boxes are displayed.}%
    \label{fig:smallpreds}
\end{figure*}

In \cref{tab:results_new}, we compare the performance of object detection methods relative to the feature extraction backbone and the detection heads.
The observations follow a pattern similar to the models trained on natural images like COCO~\cite{lin2014microsoft}, where detection performance increases with the number of model parameters.

This is illustrated in \cref{tab:results_backbone} where we compare the performance of different feature extraction backbones used in conjunction with F-RCNN~\cite{ren2015faster} detection heads and observe a significant increase in performance with stronger feature extractors.
Conversely, \cref{tab:results_arch} indicates that updating the detection heads while maintaining the same feature extraction method, also leads to an improved performance.
Notably, the recent one-stage \ac{MADet}~\cite{xie2023} method surpasses the slower two-stage F-RCNN~\cite{ren2015faster} with the same backbone. 
\ac{FCOS}~\cite{tian2019}, on the other hand, still shows a weaker performance than the two-stage algorithm. 
In \cref{tab:results_strong} we explore the extent to which detection performance can be improved by combining the best-performing detection architecture with strong SWIN-L~\cite{liu2021swin} and FocalNet-L~\cite{yang2022focal} backbones.
Additionally, we evaluate the YOLO-v8L~\cite{Jocher_YOLO_by_Ultralytics_2023} model, pre-trained on COCO~\cite{lin2014microsoft}.
Although the pre-training on COCO~\cite{lin2014microsoft} makes the YOLO model incomparable to the other models that have not been directly pre-trained for object detection, we include it in the comparison to assess how effectively this quick, ready-to-use solution can be applied to the dataset.
We hypothesize that the strong performance of the YOLO-v8L~\cite{Jocher_YOLO_by_Ultralytics_2023} model can primarily be attributed to the COCO~\cite{lin2014microsoft} pre-training targeted towards detection.

\begin{table}
\centering
     \caption{Performance of DINO detection models with strong SWIN-L~\cite{liu2021swin} and FocalNet-L~\cite{yang2022focal} backbones, and a YOLO-v8L architecture with strong COCO~\cite{lin2014microsoft} pre-training. }
     \label{tab:results_strong}
\begin{tabular}{llccc}
              \toprule
            Model & PT DS &  AP & AP$_{50}$ & AP$_{75}$\\
             \midrule
             DINO\cite{zhang2022dino} (SWIN-L~\cite{liu2021swin}) & IN21k & 19.5 & 32.2 & 20.4  \\
             DINO\cite{zhang2022dino} (FocalNet-L\cite{yang2022focal}) & IN21k & \textbf{22.6} & \textbf{36.5} & \textbf{23.5} \\
             YOLO-v8L\cite{Jocher_YOLO_by_Ultralytics_2023} & COCO & 18.5 & 29.0 & 19.8 \\
             \bottomrule
\end{tabular}
\end{table}
When combining the strong (and heavy) SWIN-L~\cite{liu2021swin} backbone with the more recent DINO~\cite{zhang2022dino} architecture, we observe a further performance boost. 
This is surpassed yet again by the most powerful model configuration, DINO~\cite{zhang2022dino} paired with a FocalNet-L~\cite{yang2022focal} backbone, exceeding the second-best performance by a large margin.
The notable increase in performance cannot be solely attributed to the larger pre-training dataset, as evidenced by the weaker performance of the DINO~\cite{zhang2022dino} and F-RCNN~\cite{ren2015faster} model configurations with a SWIN-L~\cite{liu2021swin} backbone, which have also been pre-trained with ImageNet-21k~\cite{ridnik2021imagenet}. 
Rather, it implies that the advanced attention-based context modelling applied by FocalNet~\cite{yang2022focal} is well-suited for processing complex datasets with diverse object categories. Such datasets often require taking into account varying degrees of context to achieve successful detection.

\paragraph{Impact of Object Sizes}
\begin{table*}[t]
    \centering
    \caption{Model performance evaluated on the super categories only, and on different object sizes (all numbers in percent), ``PT DS'' specifies the pretraining datasets used for model initialisation.}
    \label{tab:ablation}
        \begin{tabular}{llcccccccc}
        \toprule 
        \multirow{2}{*}{Model} & \multirow{2}{*}{PT DS}  && \multicolumn{3}{c}{Supercategory} && \multicolumn{3}{c}{Object Sizes} \\
            \cmidrule{4-6} \cmidrule{8-10} 
            &&&  AP & AP$_{50}$ & AP$_{75}$ &&  AP$_s$ & AP$_m$ & AP$_l$\\
             \midrule
             F-RCNN\cite{ren2015faster} (RN50\cite{he2016deep}) & IN1k &&  16.2 & 32.6 & 15.6 && \phantom{0}6.6 & 10.9 & 15.6 \\
             F-RCNN\cite{ren2015faster} (RN101\cite{he2016deep}) & IN1k  && 17.0 & 33.3 & 16.0 && \phantom{0}5.0 & 11.8 & 17.2 \\
             F-RCNN\cite{ren2015faster} (x101-32\cite{xie2017aggregated}) & IN1k  &&  17.5 & 35.3 & 15.1 && \phantom{0}6.6 & 11.5 & 17.7 \\
             F-RCNN\cite{ren2015faster} (x101-64\cite{xie2017aggregated}) & IN1k  && 18.9 & 37.0 & 18.0 && \phantom{0}6.8 & 12.6 & 19.5 \\
             F-RCNN\cite{ren2015faster} (SWIN-L\cite{liu2021swin}) & IN21k   &&  27.2 & 52.1 & 25.4 && 11.6 & 20.4 & 30.8 \\
             
             DINO\cite{zhang2022dino} (SWIN-L\cite{liu2021swin}) & IN21k   && 26.4 & 47.8 & 26.2 &&  11.1 & 20.2 & 35.7 \\
             DINO\cite{zhang2022dino} (FocalNet-L\cite{yang2022focal}) & IN21k  && \textbf{32.5} & \textbf{57.8} & \textbf{32.3} &&  \textbf{12.2} & \textbf{24.3} & \textbf{40.3}\\
             YOLO-v8L\cite{Jocher_YOLO_by_Ultralytics_2023} & COCO && 26.5 & 43.6 & 27.4 &&  \textbf{12.2} & 21.3 & 30.1 \\
             MADet\cite{xie2023} (RN50\cite{he2016deep}) & IN1k && 15.3 & 31.2 & 13.4 && 
             5.3 & 11.3 & 15.3  \\
             FCOS\cite{tian2019} (RN50\cite{he2016deep}) & IN1k && 14.4 & 27.4 & 14.1 &&
             5.6 & 8.8 & 13.2\\ 
             \bottomrule
        \end{tabular}
\end{table*}

One of the challenging properties of the ODOR dataset is the number of small object instances.
The size-relative performances of all models are listed in \cref{tab:ablation}, as measured by the canonical COCO~\cite{lin2014microsoft} definition of object sizes (small: area $<32^2$ px, medium: $32^2$ px $<$ area $<$ $96^2$ px, large: area $>$ $96^2$px).
Interestingly, while YOLO-v8~\cite{Jocher_YOLO_by_Ultralytics_2023} ranks third in the overall score, it is on par with the best-performing model on small objects. %

\Cref{fig:smallpreds} provides a qualitative example where small object predictions generated by YOLO-v8~\cite{Jocher_YOLO_by_Ultralytics_2023} appear equally accurate as those generated by the otherwise stronger DINO~\cite{zhang2022dino} models with large SWIN~\cite{liu2021swin} and FocalNet~\cite{yang2022focal} backbones.
While it might seem natural to attribute this to its anchorless region proposal method -- widely regarded as beneficial for detecting small objects~\cite{tian2019,xie2023} -- this hypothesis is not supported by the performance of the similarly anchorless FCOS~\cite{tian2019} model, whose performance on small objects aligns with its overall performance measures. 
Apparently, an anchorless region proposal method alone does not suffice to achieve strong small-object detection results; effective pre-training on a dedicated detection dataset is essential as well.

\paragraph{Impact of Supercategories}
In addition to the object sizes, the fine granularity of the ODOR classes poses a further challenge. 
To quantify the impact of a detailed class hierarchy, we apply a supercategory-based evaluation protocol where each detection is considered a true positive if it has the same supercategory as the respective ground truth box. 
The corresponding AP values reported in \cref{tab:ablation} exhibit a consistent \SI{35}{\percent} to \SI{60}{\percent} increase compared to the evaluation of the complete set of categories.
These results highlight the crucial role of a complex set of detection categories in designing a challenging benchmark for advanced detection methods.

\begin{figure*}
\begin{subfigure}{.5\columnwidth}
    \centering
    \includegraphics[width=.9\columnwidth]{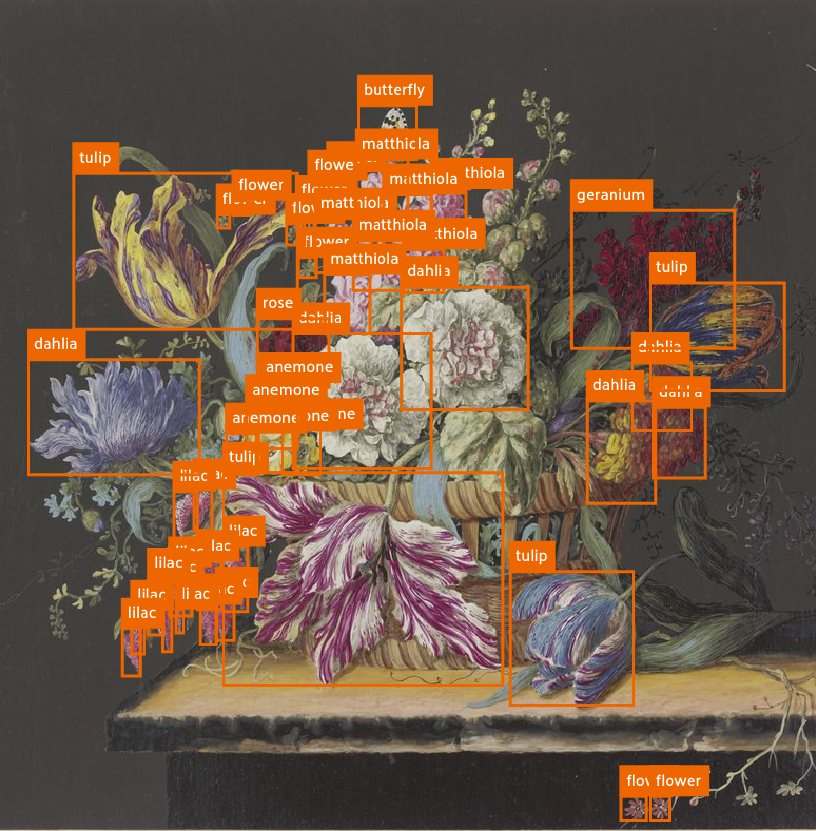}
    \caption{Ground Truth}
    \label{fig:flowers_gt}
\end{subfigure}~%
\begin{subfigure}{.5\columnwidth}
    \centering
    \includegraphics[width=.9\columnwidth]{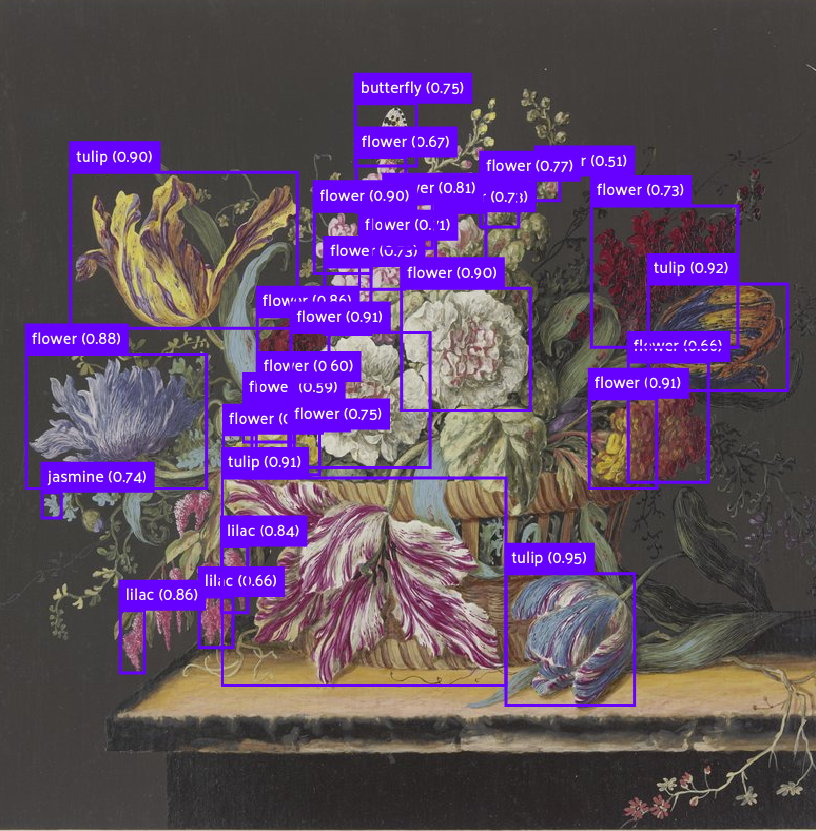}
    \caption{DINO\cite{zhang2022dino}~(FocalNet-L\cite{yang2022focal})}
    \label{fig:flowers_dino}
\end{subfigure}
\begin{subfigure}{.5\columnwidth}
    \centering
    \includegraphics[width=.9\columnwidth]{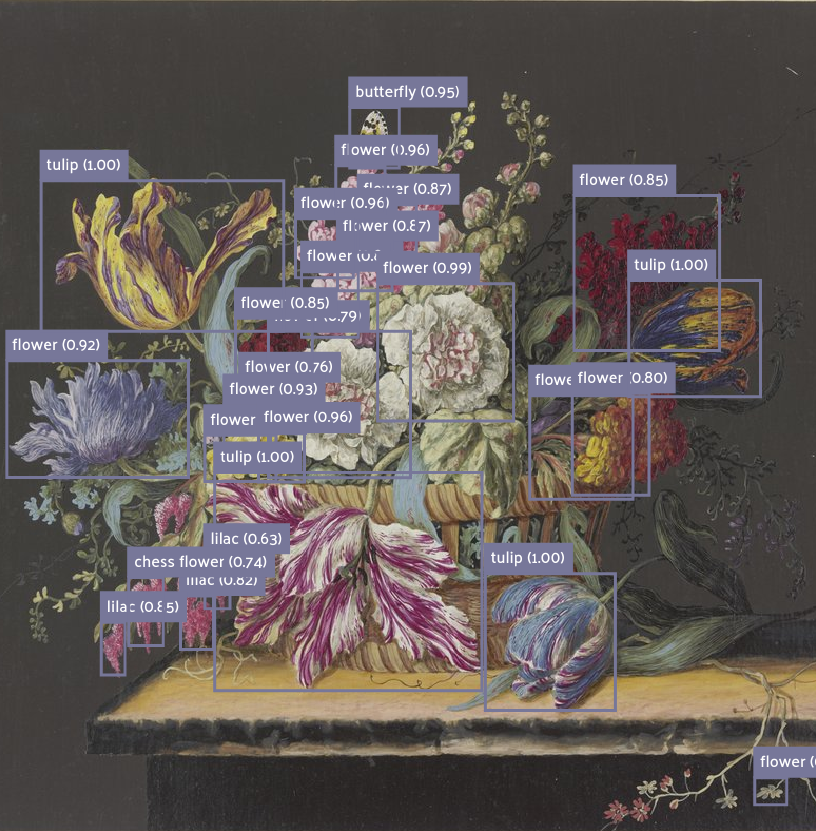}
    \caption{FRCNN\cite{ren2015faster}~(SWIN-L\cite{liu2021swin})}
    \label{fig:flowers_frcnn}
\end{subfigure}~%
\begin{subfigure}{.5\columnwidth}
    \centering
    \includegraphics[width=.9\columnwidth]{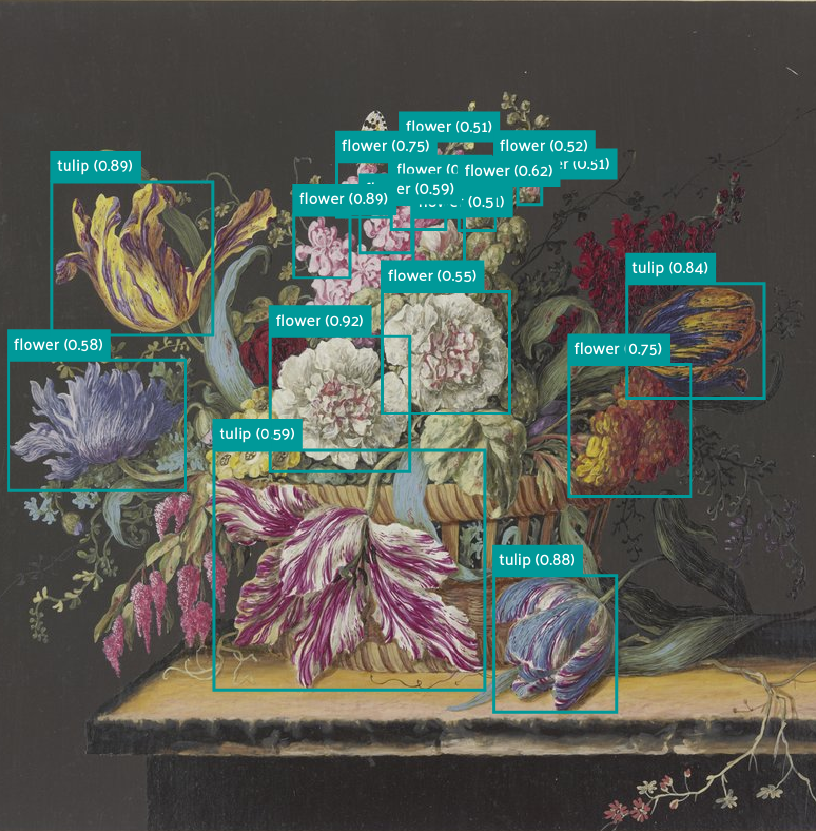}
    \caption{YOLO-v8L\cite{Jocher_YOLO_by_Ultralytics_2023}\phantom{0000000000}}
    \label{fig:flowers_yolo}
\end{subfigure}
\bigskip

\begin{subfigure}{\textwidth}
    \centering
    \begin{tabular}{lcc}
    \toprule
        Model & AP & AP~(supercategory)  \\
        \midrule
        (b) DINO\cite{zhang2022dino}~(FocalNet-L\cite{yang2022focal}) & \textbf{28.8} & 53.2 \\
        (c) FRCNN\cite{ren2015faster}~(SWIN-L\cite{liu2021swin}) & 26.7 & 32.9 \\
        (d) YOLO-v8L\cite{Jocher_YOLO_by_Ultralytics_2023} & 25.7 & \textbf{56.6} \\
        \bottomrule
    \end{tabular}
    \caption{Comparison of model performance for normal category vs. supercategory for this image.}
    \label{tab:singlemap}
\end{subfigure}
    \caption{\subref{fig:flowers_gt} A bouquet of various flower species; \subref{fig:flowers_dino}--\subref{fig:flowers_yolo} different model predictions. For clarity, only confidence scores above $0.5$ are displayed. 
    All models achieve decent localisation capabilities for the separate blossoms but fail to assign the correct flower species reliably.
    This reflects in the AP scores evaluated on the single image; \subref{tab:singlemap}
    when evaluated on the supercategories, the models consistently exhibit a significant increase in accuracy.
}
    
    \label{fig:bouquet}
\end{figure*}

\paragraph{Class-Wise Performance}
When comparing the average detection precision per class, we find a large variety in model performance.
\begin{table}[]
    \centering
    \caption{Sorted class-wise AP scores of the best performing DINO~\cite{zhang2022dino} model, the number of respective training instances and average object sizes of all test set instances. }
    \label{tab:class_map}
    \begin{tabular}{lccc}
    \toprule
        Class  &  AP  & Train Instances & Avg. Size (px.)\\
        \midrule
        Lizard & 67.4 & \phantom{00}31 & \num{22355} \\
        Cauliflower & 65.0 & \phantom{00}17 & \num{24383} \\
        Teapot & 63.5 & \phantom{0}200 & \num{29644}\\
        Jan Steen Jug & 63.0 & \phantom{00}37  &\num{40234}\\
        Censer & 60.8 & \phantom{00}52 & \phantom{0}\num{7204} \\[4pt]
        \hspace{1em}\smash{\vdots} &  \hspace{1em}\smash{\vdots} & \hspace{1.5em}\smash{\vdots} & \phantom{11111}\smash{\vdots}\\
        Grapes & 13.8 & 1576 & \phantom{11}\num{7086}\\[4pt]
        \hspace{1em}\smash{\vdots} &  \hspace{1em}\smash{\vdots} & \hspace{1.5em}\smash{\vdots} & \phantom{11111}\smash{\vdots}\\
        Heliotrope & \phantom{11}0 & \phantom{0}102 & \phantom{111}\num{440} \\
        Hyacinth & \phantom{11}0 & \phantom{00}25 & \num{30788} \\
        \bottomrule
    \end{tabular}
\end{table}
\Cref{tab:class_map} lists exemplary per-class AP values for the strongest DINO~\cite{zhang2022dino} model and relates them to their respective number of instances in the training set. 
Although having less than 40 training samples, ``Jan Steen Jug'' and ``Lizard'' both rank among the best-detected classes whereas the more frequent ``Heliotrope'' has a very low detection rate. 
Across all classes, we observe only a moderate correlation between the number of training instances and the class-wise AP with a spearman correlation coefficient of 0.38.
The long-tail class distribution does not appear to be a sufficient explanation for the observed differences in class-wise detection performance. 
Another factor that may be contributing is the average size of objects in each class in the test set.
However, the object sizes and the detection performance are only moderately correlated as well ($\rho = 0.43$).
Accordingly, ``Censer'' ranks among the categories with the highest class-wise detection rates in spite of a small average instance size. 
Conversely, ``Hyacinth'' instances can not reliably detected by the model although their average instance size is large.
We conclude that there seem to be class-specific properties that render their successful recognition more or less difficult. 

\begin{figure}
\begin{subfigure}[t]{.49\columnwidth}
    \centering
    \includegraphics[width=\textwidth]{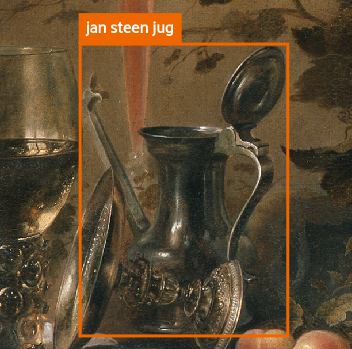}
    \caption{}
    \label{fig:jansteenjug}
\end{subfigure}~\begin{subfigure}[t]{.49\columnwidth}
    \centering
    \includegraphics[width=\textwidth]{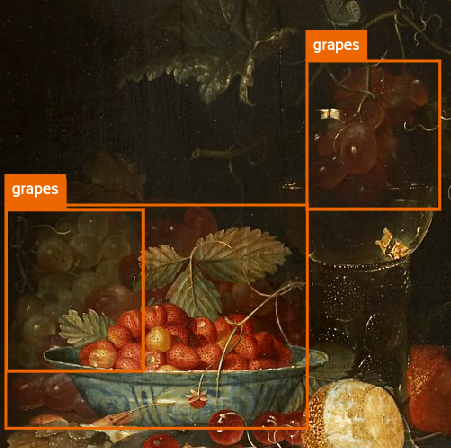}
    \caption{}
    \label{fig:grapes}
\end{subfigure}
    \caption[]{Jan Steen Jug (left) and occluding bunches of grape, partly blending into the background (right). 
    }
    \label{fig:jansteengrapes}
\end{figure}

Consider the depiction of a Jan Steen Jug in \cref{fig:jansteengrapes} (left) with its distinctive visual features such as the long spout and its characteristic surface texture.
Additionally, these jugs usually appear in similar contexts, isolated from other objects.
The grapes depicted in \cref{fig:jansteengrapes} (right) on the other hand illustrate the challenges associated with their recognition: 
Grapes usually appear in bunches, heavily occluding each other, and sometimes blending into the background. 
Although they are the second most common object in the training set with more than 1500 instances, their class-wise AP is considerably lower than the average over all classes (\SI{22.6}{\percent}).
These observations highlight the crucial role of (1) a large share of small objects, and (2) a detailed set of challenging object categories.
These factors are important to benchmark generalisable detection algorithms meeting real-world requirements.

%% file: limitations.tex
\paragraph{Image Licenses}
Some images do not possess licenses that permit redistribution.
In response to this constraint, we are unable to upload physical copies of the images.
Instead, we provide a list of download links, together with the image meta-data and license information to re-download from the respective source collection and a download script to conveniently download the images.
While we took care to always supply permanent links, as opposed to less reliable query-based download links, we cannot ensure the availability of the image resources in the same manner as if we were hosting the images ourselves.

\paragraph{Dataset Size \& Applicability}
Compared to expansive photographic datasets like COCO~\cite{lin2014microsoft}, OpenImages~\cite{kuznetsova2020open}, or Objects365~\cite{shao2019objects365}, the dataset size in terms of annotations and images is notably limited. 
This limited scale restricts its broader applicability as a training resource, particularly beyond the artistic domain. 

Nevertheless, the dataset's challenging properties make it a valuable resource for assessing domain transfer capabilities and robustness in detecting peripheral, occluding, or small objects.
Within the artistic domain, it is one of the few datasets that offer instance-level annotations.
It surpasses numerous comparable datasets like PeopleArt~\cite{westlake2016detecting}, IconArt~\cite{gonthier2018weakly}, PoPArt~\cite{schneider2023poses}, and SniffyArt~\cite{zinnen2023sniffy} but falls short in terms of the number of images and annotations when compared to the large DEArt~\cite{reshetnikov2022deart} and Human-Art~\cite{ju2023human} datasets. 
However, what sets the ODOR dataset apart is not so much the number of categories or annotations but rather its complexity as measured in instance density and class granularity. 
This complexity renders it indispensable for a wide range of applications requiring fine-grained classification, whether as a source for pre-training or direct application.

\paragraph{Historic and Artistic Ambiguity}
Historical objects are subject to changes in appearance, usage, and naming conventions. 
Annotating objects within a rigid classification scheme assumes the existence of a taxonomy of distinct and immutable categories, which fails to consider the dynamic nature of historical change and unresolved borderline cases. 
While, in theory, this is also the case with modern photographic datasets, the problem becomes especially evident when dealing with historical images spanning centuries of development and transformation. 
Furthermore, in some instances, the artists may have exercised their artistic freedom and abstraction capabilities to depict objects that defy clear categorisation, deliberately introducing ambiguity or uncertainty.
We compensate for this ambiguity by consulting experts from cultural and art history and conducting multiple rounds of corrections performed by two art history experts. 
Despite these efforts, an inherent level of ambiguity persists due to the historical nature of these images, which cannot be entirely eliminated or resolved.

%% file: image_credits.tex
\label{credits}
\scriptsize
    \begin{tabularx}{\textwidth}{lp{.82\linewidth}}
        \toprule
        Image & Credits \\
        \midrule
        \Cref{fig:examplesa} & Detail from: \textit{Still Life with Bouquet of Flowers}. Jan Brueghel the Elder. 1610 -- 1625. Oil on copper. Städel Museum, Frankfurt am Main. \url{https://www.staedelmuseum.de/go/ds/540}. \\
        \Cref{fig:examplesb} & Detail from: \textit{Village in the Snow}. David Teniers (II). 1625 -- 1690. Oil on panel. RKD -- Netherlands Institute for Art History, \href{https://rkd.nl/images/290572}{RKDImages(290572)}. \\
        \Cref{fig:examplesc} & Detail from: \textit{Man Smoking and Holding an Empty Wine Glass}. Jan Miense Molenaer. 1632 -- 1634. Oil on oak. Städel Museum, Frankfurt am Main. \url{https://www.staedelmuseum.de/go/ds/442}.\\
        \Cref{fig:description} & \textit{Stillleben mit bauchiger Flasche}. Anne Vallayer Coster. c. 1770. Staatliche Museen zu Berlin, Gemäldegalerie / Jörg P. Anders. \url{https://id.smb.museum/object/863975}.
\\
        \Cref{fig:metadata} & \textit{The parable of the rich man and poor Lazarus}. Lodewijk Toeput. 1565 -- 1605. Oil on canvas. \href{https://rkd.nl/images/108569}{RKDImages(108569)}.\\
        \Cref{fig:occla} & \textit{The Milkmaid}. Johannes Vermeer. 1657 -- 1658. Rijksmuseum Amsterdam. \url{https://www.rijksmuseum.nl/en/collection/SK-A-2344.}
\\
        \Cref{fig:occlb} & \textit{Stillleben}. August Rieper. 1899. Staatliche Museen zu Berlin, Nationalgalerie / Andreas Kilger. \url{https://id.smb.museum/object/967590/stilleben}.
 \\
        \Cref{fig:occla} &   \textit{Blumenstrauß}. Jan Brueghel the Elder. 1620. Gemäldegalerie, Staatliche Museen zu Berlin / Jörg P. Anders (CC-BY-SA). \url{https://id.smb.museum/object/865413}.\\
        \Cref{fig:occlb} &     \textit{Still life with a lobster, glasswork, bread, cheese, and parrots}. Artus Claessens. 1615 -- 1644. RKD -- Netherlands Institute for Art History, \href{https://rkd.nl/images/16311}{RKDImages(16311)}.\\
        \Cref{fig:yolov8preds} &
        Detail from \textit{Still life with a lobster, glasswork, bread, cheese, and parrots}. Artus Claessens. 1615 -- 1644. RKD -- Netherlands Institute for Art History, \href{https://rkd.nl/images/16311}{RKDImages(16311)}. \\
        \Cref{fig:bouquet} & Detail from
    \textit{Blumenkorb mit Malven, Levkojen Primeln, Tulpen und anderen Blumen auf einem Tisch}. After Barbara Regina Dietzsch. c. 1800. Städel Museum, Frankfurt am Main. \url{https://www.staedelmuseum.de/go/ds/5469z}.  \\
        \Cref{fig:jansteenjug} &     \textit{Still Life with Vessels and Fruits}. Pieter Claesz. 1642 -- 1650. Staatliche Museen zu Berlin, Gemäldegalerie / Volker-H. Schneider. \url{https://id.smb.museum/object/967590}. \\
        \Cref{fig:grapes} &     \textit{Still Life with Fruit and a Chinese Porcelain Bowl with Strawberries}. Harmen Loading. 17th Century. Städel Museum, Frankfurt am Main. \url{https://id.smb.museum/object/870390}. \\
        \bottomrule
    \end{tabularx}

%% file: training_details.tex
This section outlines the experimental settings for the baseline methods trained on the ODOR dataset.
We used established frameworks specific to each detector architecture:
Faster R-CNN was trained using the mmdetection framework\footnote{\url{https://github.com/open-mmlab/mmdetection}},
DINO via detrex\footnote{\url{https://github.com/IDEA-Research/detrex}}, 
and YOLOv8 using ultralytics\footnote{\url{https://github.com/ultralytics/ultralytics}}, with each framework employing its default data augmentation techniques.
Note that Faster R-CNN and DINO columns subsume multiple model variants with varying backbones, yet maintaining identical hyperparameter settings.
All models were trained for 50 epochs, employing the default hyperparameters proposed by the algorithm implementation in the respective frameworks. 
The specific configurations are listed in \cref{tab:hyperparams}.

\begin{table}[H]
    \centering
    \begin{tabular}{llllll}
    \toprule
         & F-RCNN & DINO & YOLO-v8 & FCOS & MADet\\
    \midrule
        input size & 800 & 800 & 640 & 800 & 800 \\
        optimizer & SGD & AdamW & auto\textsuperscript{\textdagger} & SGD & SGD\\
        base lr & 0.02 & \num{1e-4} & 0.01 & \num{1e-4} & \num{1e-3}\\
        weight decay & \num{1e-4} & \num{1e-4} & \num{5e-4} & \num{1e-4} & \num{1e-4} \\  
        momentum & 0.9 & 0.9 & 0.937 & 0.9 & 0.9\\
        batch size & 4 & 4 & 16 & 2 & 8\\
        training epochs & 50 & 50 & 50 & 50 & 50\\ 
        lr scheduler & step(16,19) & linear(36,30) & constant & step(16,19) & step(16,19)\\
        lr gamma & 0.1 & 0.1 & - & 0.1 & 0.1\\
        
    \bottomrule
    \end{tabular}
    \caption{Hyperparameters for training baseline models on the ODOR dataset. \textsuperscript{\textdagger}uses AdamW initially, then switches to SGD after \num{10000} iterations.}
    \label{tab:hyperparams}
\end{table}

%% file: authorbios.tex
\pagestyle{empty}

\emph{Mathias Zinnen} (orcid: 0000-0003-4366-5216) studied philosophy and computer science in Berlin and Erlangen. Currently pursuing a PhD in FAU's Pattern Recognition Lab, his research centers on computer vision in arts. His main academic interests cover object detection, pose estimation, and multimodal methods, along with innovative work on visual representations of smell.

\emph{Prathmesh Madhu} (orcid: 0000-0003-2707-415X) is a technical lead at Infocusp Innovations, Ahmedabad. He holds a PhD from FAU, specializing in digital humanities, applying advanced computer vision and machine learning to understand artworks. With 6+ years of experience, he's also passionate about mathematics and led projects in CV and NLP as a Machine Learning Engineer before his PhD.

\emph{Azhar Hussian} is pursuing a master's in Artificial Intelligence at FAU. He works as a student assistant at FAU's Pattern Recognition Lab until December 31, 2023, with over 2 years of prior deep learning experience.

\emph{Hang Tran} is pursuing a master's in Art History with a focus on digital art history at FAU. She submitted her thesis on tobacco and smoke representations in Dutch art. From July 2021 until the end of 2023 she works as a student assistant at FAU's Pattern Recognition Lab. 

\emph{Ali Hürriyetoğlu} (orcid: 0000-0003-3003-1783) works on historical multilingual text processing. After his dissertation focusing on social media information extraction, he was a postdoc at Koc University. He has experience in industry, government, and academia and currently researches text processing system robustness. Since 2019, he organizes events on socio-political extraction and multimodal data processing.

\emph{Inger Leemans} (orcid: 0000-0003-1640-4109) is a Professor of Cultural History at Vrije Universiteit Amsterdam and PI of NL-Lab, a Dutch Culture and Identity research group at the Royal Netherlands Academy of Arts and Sciences. Her work explores early modern cultural history, emotion and sensory history, and digital humanities. Leemans is member of the board of the American Historical Review and the Scientific board of the European Joint Programming Initiative on Cultural Heritage. As PI of the Horizon2020 project Odeuropa she leads an international research and development project on Olfactory Heritage and Sensory Mining.

\emph{Peter Bell} (orcid: 0000-0003-4415-7408) earned a PhD in Art History focusing on Greeks in Italian art. Bell was a Junior Professor for Digital Humanities at the FAU from 2017 to 2021 and now holds a professorship at the University of Marburg, teaching Art History and Digital Humanities. He researches on computer vision in art and cultural heritage. 

\emph{Andreas Maier} (orcid: 0000-0002-9550-5284) is a Professor of computer science specializing in medical signal processing. He worked at Stanford University and Siemens Healthcare, managing CT reconstruction projects. He has been the head of the Pattern Recognition Lab at FAU since 2015. He researches medical imaging, audio processing, digital humanities, and interpretable machine learning.

\emph{Vincent Christlein} (orcid: 0000-0003-0455-3799) received his Computer Science PhD from FAU in 2018, focusing on automatic handwriting analysis. He's a research associate at FAU's Pattern Recognition Lab, heading the Computer Vision group, researching various topics including environmental projects and computational humanities like document and art analysis.

%% file: odor_arxiv.bbl
\begin{thebibliography}{10}
\expandafter\ifx\csname url\endcsname\relax
  \def\url#1{\texttt{#1}}\fi
\expandafter\ifx\csname urlprefix\endcsname\relax\def\urlprefix{URL }\fi
\expandafter\ifx\csname href\endcsname\relax
  \def\href#1#2{#2} \def\path#1{#1}\fi

\bibitem{cai2015cross}
H.~Cai, Q.~Wu, T.~Corradi, P.~Hall, {The Cross-Depiction Problem: Computer Vision Algorithms for Recognising Objects in Artwork and in Photographs}, arXiv preprint arXiv:1505.00110 (2015).

\bibitem{madhu2022enhancing}
P.~Madhu, A.~Villar-Corrales, R.~Kosti, T.~Bendschus, C.~Reinhardt, P.~Bell, A.~Maier, V.~Christlein, {Enhancing Human Pose Estimation in Ancient Vase Paintings via Perceptually-grounded Style Transfer Learning}, ACM Journal on Computing and Cultural Heritage 16~(1) (2022) 1--17.

\bibitem{strezoski2018omniart}
G.~Strezoski, M.~Worring, {OmniArt}: {A Large-scale Artistic Benchmark}, ACM Transactions on Multimedia Computing, Communications, and Applications (TOMM) 14~(4) (2018) 1--21.

\bibitem{westlake2016detecting}
N.~Westlake, H.~Cai, P.~Hall, {Detecting People in Artwork with CNNs}, in: Computer Vision--ECCV 2016 Workshops: Amsterdam, The Netherlands, October 8-10 and 15-16, 2016, Proceedings, Part I 14, Springer, 2016, pp. 825--841.

\bibitem{gonthier2018weakly}
N.~Gonthier, Y.~Gousseau, S.~Ladjal, O.~Bonfait, {Weakly Supervised Object Detection in Artworks}, in: L.~Leal-Taix{\'e}, S.~Roth (Eds.), Computer Vision -- ECCV 2018 Workshops, Springer International Publishing, Cham, 2019, pp. 692--709.

\bibitem{reshetnikov2022deart}
A.~Reshetnikov, M.-C. Marinescu, J.~M. Lopez, {DEArt}: {Dataset of European Art}, in: European Conference on Computer Vision, Springer, 2022, pp. 218--233.

\bibitem{zinnen2023sniffy}
M.~Zinnen, A.~Hussian, H.~Tran, P.~Madhu, A.~Maier, V.~Christlein, \href{https://doi.org/10.1145/3607542.3617357}{{SniffyArt: The Dataset of Smelling Persons}}, in: Proceedings of the 5th Workshop on AnalySis, Understanding and ProMotion of HeritAge Contents, SUMAC '23, Association for Computing Machinery, New York, NY, USA, 2023, p. 49–58.
\newblock \href {https://doi.org/10.1145/3607542.3617357} {\path{doi:10.1145/3607542.3617357}}.
\newline\urlprefix\url{https://doi.org/10.1145/3607542.3617357}

\bibitem{ju2023human}
X.~Ju, A.~Zeng, J.~Wang, Q.~Xu, L.~Zhang, {Human-Art}: A versatile human-centric dataset bridging natural and artificial scenes, in: Proceedings of the IEEE/CVF Conference on Computer Vision and Pattern Recognition, 2023, pp. 618--629.

\bibitem{gupta2019lvis}
A.~Gupta, P.~Dollar, R.~Girshick, {LVIS}: {A Dataset for Large Vocabulary Instance Segmentation}, in: Proceedings of the IEEE/CVF conference on computer vision and pattern recognition, 2019, pp. 5356--5364.

\bibitem{zinnen2022odor}
M.~Zinnen, P.~Madhu, R.~Kosti, P.~Bell, A.~Maier, V.~Christlein, {ODOR}: The {ICPR2022} {Odeuropa Challenge on Olfactory Object Recognition}, in: 2022 26th International Conference on Pattern Recognition (ICPR), IEEE, 2022, pp. 4989--4994.

\bibitem{zhao2023automatic}
S.~Zhao, A.~Akda{\u{g}}~Salah, A.~A. Salah, {Automatic Analysis of Human Body Representations in Western Art}, in: Computer Vision--ECCV 2022 Workshops: Tel Aviv, Israel, October 23--27, 2022, Proceedings, Part I, Springer, 2023, pp. 282--297.

\bibitem{kadish2021improving}
D.~Kadish, S.~Risi, A.~S. L{\o}vlie, {Improving Object Detection in Art Images Using Only Style Transfer}, in: 2021 International Joint Conference on Neural Networks (IJCNN), IEEE, 2021, pp. 1--8.

\bibitem{hall2015cross}
P.~Hall, H.~Cai, Q.~Wu, T.~Corradi, Cross-depiction problem: Recognition and synthesis of photographs and artwork, Computational Visual Media 1 (2015) 91--103.

\bibitem{cetinic2022understanding}
E.~Cetinic, J.~She, {Understanding and Creating Art with AI: Review and Outlook}, ACM Transactions on Multimedia Computing, Communications, and Applications (TOMM) 18~(2) (2022) 1--22.

\bibitem{lu2022data}
Y.~Lu, C.~Guo, X.~Dai, F.-Y. Wang, Data-efficient image captioning of fine art paintings via virtual-real semantic alignment training, Neurocomputing 490 (2022) 163--180.

\bibitem{sabatelli2018deep}
M.~Sabatelli, M.~Kestemont, W.~Daelemans, P.~Geurts, {Deep Transfer Learning for Art Classification Problems}, in: L.~Leal-Taix{\'e}, S.~Roth (Eds.), Computer Vision -- ECCV 2018 Workshops, Springer International Publishing, Cham, 2019, pp. 631--646.

\bibitem{gonthier2021analysis}
N.~Gonthier, Y.~Gousseau, S.~Ladjal, An analysis of the transfer learning of convolutional neural networks for artistic images, in: Pattern Recognition. ICPR International Workshops and Challenges: Virtual Event, January 10--15, 2021, Proceedings, Part III, Springer, 2021, pp. 546--561.

\bibitem{zinnen2022transfer}
M.~Zinnen, P.~Madhu, P.~Bell, A.~Maier, V.~Christlein, {Transfer Learning for Olfactory Object Detection}, in: Digital Humanities Conference, 2022, Alliance of Digital Humanities Organizations, 2022, pp. 409--413, https://arxiv.org/abs/2301.09906.

\bibitem{zhao2022big}
W.~Zhao, W.~Jiang, X.~Qiu, {Big Transfer Learning for Fine Art Classification}, Computational Intelligence and Neuroscience 2022 (2022).

\bibitem{fisherdisc}
G.~Cheng, J.~Han, P.~Zhou, D.~Xu, {Learning Rotation-Invariant and Fisher Discriminative Convolutional Neural Networks for Object Detection}, {IEEE Transactions on Image Processing} 28~(1) (2019) 265--278.
\newblock \href {https://doi.org/10.1109/TIP.2018.2867198} {\path{doi:10.1109/TIP.2018.2867198}}.

\bibitem{russakovsky2015imagenet}
O.~Russakovsky, J.~Deng, H.~Su, J.~Krause, S.~Satheesh, S.~Ma, Z.~Huang, A.~Karpathy, A.~Khosla, M.~Bernstein, et~al., {Imagenet Large Scale Visual Recognition Challenge}, {International Journal of Computer Vision} 115 (2015) 211--252.

\bibitem{lin2014microsoft}
T.-Y. Lin, M.~Maire, S.~Belongie, J.~Hays, P.~Perona, D.~Ramanan, P.~Doll{\'a}r, C.~L. Zitnick, Microsoft {COCO}: Common objects in context, in: Computer Vision--ECCV 2014: 13th European Conference, Zurich, Switzerland, September 6-12, 2014, Proceedings, Part V 13, Springer, 2014, pp. 740--755.

\bibitem{kuznetsova2020open}
A.~Kuznetsova, H.~Rom, N.~Alldrin, J.~Uijlings, I.~Krasin, J.~Pont-Tuset, S.~Kamali, S.~Popov, M.~Malloci, A.~Kolesnikov, et~al., {The Open Images Dataset v4: Unified image classification, object detection, and visual relationship detection at scale}, International Journal of Computer Vision 128~(7) (2020) 1956--1981.

\bibitem{cheng2020remote}
G.~Cheng, X.~Xie, J.~Han, L.~Guo, G.-S. Xia, Remote sensing image scene classification meets deep learning: {Challenges}, methods, benchmarks, and opportunities, {IEEE Journal of Selected Topics in Applied Earth Observations and Remote Sensing} 13 (2020) 3735--3756.

\bibitem{cheng2021feature}
G.~Cheng, C.~Lang, M.~Wu, X.~Xie, X.~Yao, J.~Han, {Feature Enhancement Network for Object Detection in Optical Remote Sensing Images}, {Journal of Remote Sensing} (2021).

\bibitem{crowley2014state}
E.~Crowley, A.~Zisserman, {The State of the Art: Object Retrieval in Paintings using Discriminative Regions}, in: Proceedings of the British Machine Vision Conference. BMVA Press, 2014.

\bibitem{crowley2015search}
E.~J. Crowley, A.~Zisserman, {In Search of Art}, in: Computer Vision-ECCV 2014 Workshops: Zurich, Switzerland, September 6-7 and 12, 2014, Proceedings, Part I 13, Springer, 2015, pp. 54--70.

\bibitem{crowley2016art}
E.~J. Crowley, A.~Zisserman, {The Art of Detection}, in: Computer Vision--ECCV 2016 Workshops: Amsterdam, The Netherlands, October 8-10 and 15-16, 2016, Proceedings, Part I 14, Springer, 2016, pp. 721--737.

\bibitem{girshick2015fast}
R.~Girshick, Fast {R-CNN}, in: Proceedings of the IEEE international conference on computer vision, 2015, pp. 1440--1448.

\bibitem{madhu2022one}
P.~Madhu, A.~Meyer, M.~Zinnen, L.~M{\"u}hrenberg, D.~Suckow, T.~Bendschus, C.~Reinhardt, P.~Bell, U.~Verstegen, R.~Kosti, et~al., {One-shot Object Detection in Heterogeneous Artwork Datasets}, in: 2022 Eleventh International Conference on Image Processing Theory, Tools and Applications (IPTA), IEEE, 2022, pp. 1--6.

\bibitem{madhu22icc}
P.~Madhu, T.~Marquart, R.~Kosti, D.~Suckow, P.~Bell, A.~Maier, V.~Christlein, \href{https://www.sciencedirect.com/science/article/pii/S003132032200632X}{{ICC++}: Explainable feature learning for art history using image compositions}, Pattern Recognition 136 (2023) 109153.
\newblock \href {https://doi.org/https://doi.org/10.1016/j.patcog.2022.109153} {\path{doi:https://doi.org/10.1016/j.patcog.2022.109153}}.
\newline\urlprefix\url{https://www.sciencedirect.com/science/article/pii/S003132032200632X}

\bibitem{madhu2021understanding}
P.~Madhu, T.~Marquart, R.~Kosti, P.~Bell, A.~Maier, V.~Christlein, {Understanding Compositional Structures in Art Historical Images Using Pose and Gaze Priors: Towards Scene Understanding in Digital Art History}, in: Computer Vision--ECCV 2020 Workshops: Glasgow, UK, August 23--28, 2020, Proceedings, Part II, Springer, 2021, pp. 109--125.

\bibitem{bell2019ikonographie}
P.~Bell, L.~Impett, {The Choreography of the Annunciation through a Computational Eye}, Histoire de l'art 34~(87) (2021) 01--06.

\bibitem{bernasconi2022gab}
V.~Bernasconi, {GAB}-{Gestures for Artworks Browsing}, in: 27th International Conference on Intelligent User Interfaces, 2022, pp. 50--53.

\bibitem{impett2017totentanz}
L.~Impett, F.~Moretti, {Totentanz. Operationalizing Aby Warburg’s Pathosformeln}, Tech. rep., Stanford Literary Lab (2017).

\bibitem{impett2020analyzing}
L.~Impett, {Analyzing Gesture in Digital Art History}, in: The Routledge Companion to Digital Humanities and Art History, Routledge, 2020, pp. 386--407.

\bibitem{pathos}
C.~M. Becker, {Aby Warburg's Pathosformel as Methodological Paradigm}, The Journal of Art Historiography 9 (2013) 9--CB1.

\bibitem{kim2018seeing}
S.~Kim, J.~Park, J.~Bang, H.~Lee, {Seeing is Smelling: Localizing Odor-Related Objects in Images}, in: Proceedings of the 9th Augmented Human International Conference, 2018, pp. 1--9.

\bibitem{eda2023detection}
Y.~Eda, H.~Matsukura, Y.~Nozaki, M.~Sakamoto, {Detection of odor-related objects in images based on everyday odors in {Japan}}, in: Proceedings of the AAAI Spring Symposium: Socially Responsible AI for Well-being, 2023, pp. 59--60.

\bibitem{rodriguez2020image}
N.~Rodr{\'\i}guez-Ortega, {Image Processing and Computer Vision in the Field of Art History}, in: The Routledge Companion to Digital Humanities and Art History, Routledge, 2020, pp. 338--357.

\bibitem{naslund2020cultures}
A.~N{\"a}slund~Dahlgren, A.~Wasielewski, {Cultures of Digitization: A Historiographic Perspective on Digital Art History}, Visual Resources 36~(4) (2020) 339--359.

\bibitem{lang2018reflecting}
S.~Lang, B.~Ommer, {Reflecting on How Artworks Are Processed and Analyzed by Computer Vision}, in: Proceedings of the European Conference on Computer Vision (ECCV) Workshops, 2018.

\bibitem{appadurai1988social}
A.~Appadurai, {The Social Life of Things}, Tech. rep., Cambridge University Press (1988).

\bibitem{hicks2010oxford}
D.~Hicks, M.~C. Beaudry, {The Oxford handbook of material culture studies}, OUP Oxford, 2010.

\bibitem{van2021materials}
M.~J. Van~Zuijlen, H.~Lin, K.~Bala, S.~C. Pont, M.~W. Wijntjes, {Materials In Paintings} {(MIP)}: {An interdisciplinary dataset for perception, art history, and computer vision}, Plos one 16~(8) (2021) e0255109.

\bibitem{leemans2022wind}
I.~Leemans, W.~de~Vries, {Wind Trade: How the Concept of Wind Came to Embody Speculation in the Dutch Republic}, The Journal of Modern History 94~(2) (2022) 288--325.

\bibitem{everingham20062005}
M.~Everingham, A.~Zisserman, C.~K. Williams, L.~Van~Gool, M.~Allan, C.~M. Bishop, O.~Chapelle, N.~Dalal, T.~Deselaers, G.~Dork{\'o}, et~al., The 2005 {PASCAL} visual object classes challenge, in: Machine Learning Challenges. Evaluating Predictive Uncertainty, Visual Object Classification, and Recognising Tectual Entailment: First PASCAL Machine Learning Challenges Workshop, MLCW 2005, Southampton, UK, April 11-13, 2005, Revised Selected Papers, Springer, 2006, pp. 117--176.

\bibitem{madhu2023icc++}
P.~Madhu, T.~Marquart, R.~Kosti, D.~Suckow, P.~Bell, A.~Maier, V.~Christlein, Icc++: Explainable feature learning for art history using image compositions, Pattern Recognition 136 (2023) 109153.

\bibitem{garcia2018read}
N.~Garcia, G.~Vogiatzis, {How to Read Paintings: Semantic Art Understanding with Multi-Modal Retrieval}, in: Proceedings of the European Conference on Computer Vision (ECCV) Workshops, 2018.

\bibitem{garcia2020dataset}
N.~Garcia, C.~Ye, Z.~Liu, Q.~Hu, M.~Otani, C.~Chu, Y.~Nakashima, T.~Mitamura, {A Dataset and Baselines for Visual Question Answering on Art}, in: Computer Vision--ECCV 2020 Workshops: Glasgow, UK, August 23--28, 2020, Proceedings, Part II 16, Springer, 2020, pp. 92--108.

\bibitem{wilber2017bam}
M.~J. Wilber, C.~Fang, H.~Jin, A.~Hertzmann, J.~Collomosse, S.~Belongie, {BAM}! the {Behance Artistic Media} {Dataset for Recognition Beyond Photography}, in: {Proceedings of the IEEE International Conference on Computer Vision}, 2017, pp. 1202--1211.

\bibitem{glamsurvey}
A.~Wallace, D.~McCarthy, Survey of {GLAM} {Open Access Policy and Practice}, \url{https://docs.google.com/document/d/15U__Z50WCUM_OWQ9HKLvLMlkcMoCN68FLVl9OKJQ8yY/edit?usp=sharing}, accessed: 2023-02-02 (2018).

\bibitem{schneider2023poses}
S.~Schneider, R.~Vollmer, {Poses of People in Art: A Data Set for Human Pose Estimation in Digital Art History}, arXiv preprint arXiv:2301.05124 (2023).

\bibitem{gonthier_nicolas_2018_4737435}
N.~Gonthier, \href{https://doi.org/10.5281/zenodo.4737435}{{IconArt Dataset}} (Oct. 2018).
\newline\urlprefix\url{https://doi.org/10.5281/zenodo.4737435}

\bibitem{marinescu2020improving}
M.-C. Marinescu, A.~Reshetnikov, J.~M. L{\'o}pez, {Improving Object Detection in Paintings based on Time Contexts}, in: 2020 International Conference on Data Mining Workshops (ICDMW), IEEE, 2020, pp. 926--932.

\bibitem{shao2019objects365}
S.~Shao, Z.~Li, T.~Zhang, C.~Peng, G.~Yu, X.~Zhang, J.~Li, J.~Sun, {Objects365: A Large-scale, High-quality Dataset for Object Detection}, in: Proceedings of the IEEE/CVF international conference on computer vision, 2019, pp. 8430--8439.

\bibitem{couprie1978iconclass}
L.~D. Couprie, Iconclass, a device for the iconographical analysis of art objects, Museum International 30~(3-4) (1978) 194--198.

\bibitem{brandhorst2016iconclass}
H.~Brandhorst, E.~Posthumus, Iconclass: a key to collaboration in the digital humanities, in: The Routledge Companion to Medieval Iconography, Routledge, 2016, pp. 201--218.

\bibitem{lisena2022capturing}
P.~Lisena, D.~Schwabe, M.~van Erp, R.~Troncy, W.~Tullett, I.~Leemans, L.~Marx, S.~C. Ehrich, {Capturing the Semantics of Smell: The Odeuropa Data Model for Olfactory Heritage Information}, in: The Semantic Web: 19th International Conference, ESWC 2022, Hersonissos, Crete, Greece, May 29--June 2, 2022, Proceedings, Springer, 2022, pp. 387--405.

\bibitem{miller1995wordnet}
G.~A. Miller, {WordNet: a Lexical Database for English}, Communications of the ACM 38~(11) (1995) 39--41.

\bibitem{redmon2017yolo9000}
J.~Redmon, A.~Farhadi, {YOLO9000}: better, faster, stronger, in: Proceedings of the IEEE conference on computer vision and pattern recognition, 2017, pp. 7263--7271.

\bibitem{radford2021learning}
A.~Radford, J.~W. Kim, C.~Hallacy, A.~Ramesh, G.~Goh, S.~Agarwal, G.~Sastry, A.~Askell, P.~Mishkin, J.~Clark, et~al., {Learning Transferable Visual Models from Natural Language Supervision}, in: International Conference on Machine Learning, PMLR, 2021, pp. 8748--8763.

\bibitem{kamath_mdetr_2021}
A.~Kamath, M.~Singh, Y.~LeCun, I.~Misra, G.~Synnaeve, N.~Carion, \href{http://arxiv.org/abs/2104.12763}{{MDETR} -- {Modulated} {Detection} for {End}-to-{End} {Multi}-{Modal} {Understanding}}, arXiv:2104.12763 [cs]ArXiv: 2104.12763 (Apr. 2021).
\newline\urlprefix\url{http://arxiv.org/abs/2104.12763}

\bibitem{li_grounded_2022}
L.~H. Li, P.~Zhang, H.~Zhang, J.~Yang, C.~Li, Y.~Zhong, L.~Wang, L.~Yuan, L.~Zhang, J.-N. Hwang, K.-W. Chang, J.~Gao, \href{https://ieeexplore.ieee.org/document/9879567/}{Grounded {Language}-{Image} {Pre}-training}, in: 2022 {IEEE}/{CVF} {Conference} on {Computer} {Vision} and {Pattern} {Recognition} ({CVPR}), IEEE, New Orleans, LA, USA, 2022, pp. 10955--10965.
\newblock \href {https://doi.org/10.1109/CVPR52688.2022.01069} {\path{doi:10.1109/CVPR52688.2022.01069}}.
\newline\urlprefix\url{https://ieeexplore.ieee.org/document/9879567/}

\bibitem{liu_grounding_2023}
S.~Liu, Z.~Zeng, T.~Ren, F.~Li, H.~Zhang, J.~Yang, C.~Li, J.~Yang, H.~Su, J.~Zhu, L.~Zhang, \href{http://arxiv.org/abs/2303.05499}{Grounding {DINO}: {Marrying} {DINO} with {Grounded} {Pre}-{Training} for {Open}-{Set} {Object} {Detection}}, arXiv:2303.05499 [cs] (Mar. 2023).
\newline\urlprefix\url{http://arxiv.org/abs/2303.05499}

\bibitem{xie_described_2023}
C.~Xie, Z.~Zhang, Y.~Wu, F.~Zhu, R.~Zhao, S.~Liang, \href{http://arxiv.org/abs/2307.12813}{Described {Object} {Detection}: {Liberating} {Object} {Detection} with {Flexible} {Expressions}}, arXiv:2307.12813 [cs] (Oct. 2023).
\newline\urlprefix\url{http://arxiv.org/abs/2307.12813}

\bibitem{everingham2009pascal}
M.~Everingham, L.~Van~Gool, C.~K. Williams, J.~Winn, A.~Zisserman, The pascal visual object classes {(VOC)} challenge, International journal of computer vision 88 (2009) 303--308.

\bibitem{pont2015boosting}
J.~Pont-Tuset, L.~Van~Gool, Boosting object proposals: From {PASCAL to COCO}, in: Proceedings of the IEEE international conference on computer vision, 2015, pp. 1546--1554.

\bibitem{zhou2017scene}
B.~Zhou, H.~Zhao, X.~Puig, S.~Fidler, A.~Barriuso, A.~Torralba, {Scene Parsing through ADE20K Dataset}, in: Proceedings of the IEEE conference on computer vision and pattern recognition, 2017, pp. 633--641.

\bibitem{liu2021survey}
Y.~Liu, P.~Sun, N.~Wergeles, Y.~Shang, A survey and performance evaluation of deep learning methods for small object detection, Expert Systems with Applications 172 (2021) 114602.

\bibitem{van2023more}
M.~van Erp, W.~Tullett, V.~Christlein, T.~Ehrhart, A.~H{\"u}rriyeto{\u{g}}lu, I.~Leemans, P.~Lisena, S.~Menini, D.~Schwabe, S.~Tonelli, et~al., {More than the Name of the Rose: How to Make Computers Read, See, and Organize Smells}, The American Historical Review 128~(1) (2023) 335--369.

\bibitem{magnusson2013microhistory}
S.~G. Magn{\'u}sson, I.~M. Szij{\'a}rt{\'o}, {What is Microhistory?: Theory and Practice}, Routledge, 2013.

\bibitem{ali2022musti}
H.~Ali, T.~Paccosi, S.~Menini, Z.~Mathias, L.~Pasquale, A.~Kiymet, T.~Rapha{\"e}l, M.~van Erp, {MUSTI-Multimodal Understanding of Smells in Texts and Images at MediaEval 2022}, in: Proceedings of MediaEval 2022 CEUR Workshop, 2022.

\bibitem{howes2010sensual}
D.~Howes, {Sensual Relations: Engaging the Senses in Culture and Social Theory}, University of Michiga/n Press, 2010.

\bibitem{tullett2022smell}
W.~Tullett, et~al., {Smell, History, and Heritage}, The American Historical Review 127~(1) (2022) 261--309.

\bibitem{he2016deep}
K.~He, X.~Zhang, S.~Ren, J.~Sun, {Deep Residual Learning for Image Recognition}, in: Proceedings of the IEEE conference on computer vision and pattern recognition, 2016, pp. 770--778.

\bibitem{xie2017aggregated}
S.~Xie, R.~Girshick, P.~Doll{\'a}r, Z.~Tu, K.~He, {Aggregated Residual Transformations for Deep Neural Networks}, in: Proceedings of the IEEE conference on computer vision and pattern recognition, 2017, pp. 1492--1500.

\bibitem{liu2021swin}
Z.~Liu, Y.~Lin, Y.~Cao, H.~Hu, Y.~Wei, Z.~Zhang, S.~Lin, B.~Guo, {Swin Transformer: Hierarchical Vision Transformer using Shifted Windows}, in: Proceedings of the IEEE/CVF international conference on computer vision, 2021, pp. 10012--10022.

\bibitem{ridnik2021imagenet}
T.~Ridnik, E.~Ben-Baruch, A.~Noy, L.~Zelnik-Manor, {Imagenet-21k Pretraining for the Masses}, arXiv preprint arXiv:2104.10972 (2021).

\bibitem{yang2022focal}
J.~Yang, C.~Li, X.~Dai, J.~Gao, {Focal Modulation Networks}, Advances in Neural Information Processing Systems 35 (2022) 4203--4217.

\bibitem{ren2015faster}
S.~Ren, K.~He, R.~Girshick, J.~Sun, Faster {R-CNN}: {Towards Real-Time Object Detection with Region Proposal Networks}, Advances in neural information processing systems 28 (2015).

\bibitem{he2017mask}
K.~He, G.~Gkioxari, P.~Doll{\'a}r, R.~Girshick, Mask {R-CNN}, in: Proceedings of the IEEE international conference on computer vision, 2017, pp. 2961--2969.

\bibitem{lin2017feature}
T.-Y. Lin, P.~Doll{\'a}r, R.~Girshick, K.~He, B.~Hariharan, S.~Belongie, {Feature Pyramid Networks for Object Detection}, in: Proceedings of the IEEE conference on computer vision and pattern recognition, 2017, pp. 2117--2125.

\bibitem{cai2018cascade}
Z.~Cai, N.~Vasconcelos, Cascade {R-CNN}: {Delving into High Quality Object Detection}, in: Proceedings of the IEEE conference on computer vision and pattern recognition, 2018, pp. 6154--6162.

\bibitem{paperswithcode}
{Papers with Code}, {COCO} object detection leaderboard, \url{https://paperswithcode.com/sota/object-detection-on-coco}, accessed: February 19, 2023 (2021).

\bibitem{zhang2022dino}
H.~Zhang, F.~Li, S.~Liu, L.~Zhang, H.~Su, J.~Zhu, L.~Ni, H.~Shum, {DINO}: {DETR} {with Improved Denoising Anchor Boxes for End-to-End Object Detection}, in: International Conference on Learning Representations, 2022.

\bibitem{wang2022internimage}
W.~Wang, J.~Dai, Z.~Chen, Z.~Huang, Z.~Li, X.~Zhu, X.~Hu, T.~Lu, L.~Lu, H.~Li, et~al., {InternImage}: {Exploring Large-Scale Vision Foundation Models with Deformable Convolutions}, arXiv preprint arXiv:2211.05778 (2022).

\bibitem{carion2020end}
N.~Carion, F.~Massa, G.~Synnaeve, N.~Usunier, A.~Kirillov, S.~Zagoruyko, {End-to-End Object Detection with Transformers}, in: Computer Vision--ECCV 2020: 16th European Conference, Glasgow, UK, August 23--28, 2020, Proceedings, Part I 16, Springer, 2020, pp. 213--229.

\bibitem{tian2019}
Z.~Tian, C.~Shen, H.~Chen, T.~He, \href{https://ieeexplore.ieee.org/document/9010746/}{{FCOS}: {Fully} {Convolutional} {One}-{Stage} {Object} {Detection}}, in: 2019 {IEEE}/{CVF} {International} {Conference} on {Computer} {Vision} ({ICCV}), IEEE, Seoul, Korea (South), 2019, pp. 9626--9635.
\newblock \href {https://doi.org/10.1109/ICCV.2019.00972} {\path{doi:10.1109/ICCV.2019.00972}}.
\newline\urlprefix\url{https://ieeexplore.ieee.org/document/9010746/}

\bibitem{xie2023}
X.~Xie, C.~Lang, S.~Miao, G.~Cheng, K.~Li, J.~Han, \href{https://ieeexplore.ieee.org/document/10265160/}{Mutual-{Assistance} {Learning} for {Object} {Detection}}, IEEE Transactions on Pattern Analysis and Machine Intelligence 45~(12) (2023) 15171--15184.
\newblock \href {https://doi.org/10.1109/TPAMI.2023.3319634} {\path{doi:10.1109/TPAMI.2023.3319634}}.
\newline\urlprefix\url{https://ieeexplore.ieee.org/document/10265160/}

\bibitem{Jocher_YOLO_by_Ultralytics_2023}
G.~Jocher, A.~Chaurasia, J.~Qiu, {YOLO by Ultralytics}, \url{{https://github.com/ultralytics/ultralytics}} (1 2023).

\end{thebibliography}
